\documentclass{article}

\usepackage{arxiv}

\usepackage[utf8]{inputenc}
\usepackage[T1]{fontenc}
\usepackage{hyperref}
\usepackage{url}
\usepackage{nicefrac}
\usepackage{microtype}
\usepackage{lipsum}

\usepackage{graphicx}
\usepackage{latexsym}

\usepackage{amsmath,amssymb,amsthm,amsfonts}
\usepackage{algorithm}
\usepackage{algorithmic}

\usepackage{bm}
\usepackage{xcolor}
\usepackage{colortbl}
\usepackage[caption=false]{subfig}
\usepackage{multirow}

\usepackage[square,numbers,sort]{natbib}
\newtheorem{definition}{Definition}

\DeclareMathOperator*{\argmin}{arg\,min}
\DeclareMathOperator*{\argmax}{arg\,max}

\title{Spectral Normalized-Cut Graph Partitioning with Fairness Constraints}

\author{
  Jia Li, Yanhao Wang \\
  School of Data Science and Engineering\\
  East China Normal University\\
  Shanghai, China\\
  \texttt{jiali@stu.ecnu.edu.cn, yhwang@dase.ecnu.edu.cn} \\
  \And
  Arpit Merchant \\
  Department of Computer Science\\
  University of Helsinki\\
  Helsinki, Finland\\
  \texttt{arpit.merchant@helsinki.fi} \\
}

\begin{document}

\maketitle

\begin{abstract}
Normalized-cut graph partitioning aims to divide the set of nodes in a graph into $k$ disjoint clusters to minimize the fraction of the total edges between any cluster and all other clusters. In this paper, we consider a fair variant of the partitioning problem wherein nodes are characterized by a categorical sensitive attribute (e.g., \emph{gender} or \emph{race}) indicating membership to different demographic groups. Our goal is to ensure that each group is approximately proportionally represented in each cluster while minimizing the normalized cut value. To resolve this problem, we propose a two-phase spectral algorithm called FNM. In the first phase, we add an augmented Lagrangian term based on our fairness criteria to the objective function for obtaining a fairer spectral node embedding. Then, in the second phase, we design a rounding scheme to produce $k$ clusters from the fair embedding that effectively trades off fairness and partition quality. Through comprehensive experiments on nine benchmark datasets, we demonstrate the superior performance of FNM compared with three baseline methods.
\end{abstract}

\section{Introduction}

Machine learning algorithms are widely used to make decisions that can directly affect people's lives in various domains, including banking \cite{FriedlerSVCHR19}, healthcare \cite{ChenPRJFG21}, education \cite{Mathioudakis0BC20}, and criminal justice \cite{BerkHJKR21}, to name a few.
However, a large body of work \cite{ChouldechovaR20, MehrabiMSLG21} has indicated that these algorithms, if left unchecked, often present discriminatory outcomes for particular demographic groups.
To address such concerns, recent studies have incorporated different notions of fairness into unsupervised learning problems \cite{Chierichetti0LV17, ChenFLM19, KleindessnerAM19, MahabadiV20}.
In particular, Chierichetti \emph{et al.} \cite{Chierichetti0LV17} pioneered \emph{fair clustering} for a set of points represented as vectors in Euclidean space and further characterized by a categorical sensitive attribute (e.g., \emph{gender} or \emph{race}) indicating membership in a demographic group (e.g., \emph{female} or \emph{Asian}).
In addition to minimizing the typical clustering objective, the problem also requires that the proportion of each demographic group within each cluster is roughly the same as its proportion in the dataset population (referred to as \emph{proportional fairness}).
Beyond clustering, however, fairness in the partitioning of graphs is relatively under-explored despite its broad applications to community detection \cite{Newman06, ChiangWD12, LierdeCC20} and computer vision \cite{ChewC15, ZhangFWWL0Y19}.

In this paper, we aim to fill this gap by defining a fair version of the normalized-cut graph partitioning problem \cite{ShiM00, YuS03}.
Informally, for a graph $G = (V, E)$, the original partitioning objective is to divide the set $V$ of nodes into $k$ disjoint clusters such that the fractions of inter-cluster edges are minimized while the fractions of intra-cluster edges are maximized.
Such an objective is measured by the normalized cut (Ncut) value.
In our fair variant, we further assume that each node belongs to one of $m$ sensitive groups and consider the notion of \emph{range-based proportional fairness} \cite{BeraCFN19} generalized from that in \cite{Chierichetti0LV17}.
Specifically, this requires that, in each of the $k$ clusters, the proportion of nodes of any group $c \in \{1, 2, \ldots, m\}$ is at least $\beta_c$ (lower bound) and at most $\alpha_c$ (upper bound) for two parameters $\beta_c, \alpha_c \in [0, 1]$.
Our overall objective, thus, is to produce a $k$-partition that minimizes the Ncut value while also satisfying the above fairness constraint.

To the best of our knowledge, the most relevant algorithms to our problem are those for spectral clustering with group fairness constraints \cite{KleindessnerSAM19,pmlr-v206-wang23h} due to the inherent connection between spectral clustering and normalized-cut graph partitioning.
But those algorithms suffer from two key limitations when applied to graph partitioning.
First, although they incorporate a special case of the range-based fairness constraint with $\alpha_c = \beta_c, \forall c \in \{1, 2, \ldots, m\}$ (i.e., the original \emph{proportional fairness} in \cite{Chierichetti0LV17}) into spectral node embedding, they still consider running standard $k$-means \cite{Lloyd82} on node vectors to obtain a $k$-partition. Consequently, they cannot guarantee how close the partitioning is to satisfying the (original or range-based) fairness constraint. Second, they do not provide any tunable trade-off between fairness and partition quality.

\smallskip\noindent\textbf{Our Contributions.}
In this paper, we propose a novel algorithmic framework for fair normalized-cut graph partitioning that addresses the above two limitations.
That is, we parameterize the desired level of range-based proportional fairness as a constraint to be satisfied and naturally trade off the Ncut value (i.e., \emph{quality}) and proportionality (i.e., \emph{fairness}) of the partitioning.
Similar to \cite{ShiM00, YuS03}, we transform the problem of minimizing the Ncut value of a graph into an equivalent trace minimization problem on its Laplacian matrix. 
Generally, our algorithm, which we refer to as FNM, comprises two phases.
In the first phase, we relax the original integer trace minimization problem to allow fractional memberships, add an augmented Lagrangian term \cite{Wright99} based on our fairness criteria to the objective function of the relaxed problem, and use the OptStiefelGBB method \cite{WenY13} to obtain a fairer embedding from which the partitioning found is closer to being fair.
Then, in the second phase, we apply a novel rounding scheme, adapted from Lloyd's $k$-means clustering algorithm \cite{Lloyd82}, to generate a fair partitioning from node vectors in the embedding space.
Specifically, we initialize the cluster centers, alternately assign vectors fairly to clusters, and update the centers for a fixed number of iterations or until the stopping condition is met.
Each assignment step first solves a linear program to produce $k$ nearly-fair clusters and then performs reassignments to construct strictly-fair clusters without significantly reducing partition quality.

Finally, we evaluate the performance of our FNM algorithm on nine benchmarking datasets ranging in size from 155 to 1.6M nodes along three metrics -- partition quality, fairness, and time efficiency, compared to three competitive baselines and ten variants for ablation study.
Our key findings are summarized below:
\emph{i}) FNM offers improved trade-offs between fairness and partition quality compared to the three baselines while scaling effectively to million-node-sized graphs;
\emph{ii}) our proposed fair embedding and rounding algorithms independently and jointly improve the quality of graph partitions with range-based proportional fairness constraints compared to general node embeddings and rounding schemes.

\subsection{Other Related Works}

\noindent\textbf{Min-Cut Graph Partitioning.}
Partitioning nodes in a graph into disjoint subsets to minimize an objective function, such as \emph{ratio cut} \cite{HagenK92}, \emph{normalized cut} \cite{ShiM00}, and \emph{Cheeger cut} \cite{Cheeger71}, is a fundamental combinatorial optimization problem.
Since all those graph partitioning problems are NP-hard \cite{HagenK92, ShiM00}, different heuristic algorithms were designed for them, among which spectral methods \cite{HagenK92, ShiM00, NgJW01, HanXN17} have attracted the most attention.
The basic idea of spectral methods is to relax the original integer minimization problems into continuous optimization problems, which are solved by computing the eigenvectors of the Laplacian matrix, and to find the partitioning using $k$-means or an alternative rounding method.
Other relaxation-based methods \cite{JiaDDX16, ChenHNHYH18, LiNL18} improved the efficiency over original spectral methods by avoiding eigendecomposition.
However, none of them incorporate the notion of \emph{fairness} into graph partitioning problems.

\smallskip\noindent\textbf{Fair Clustering.}
There has been rich literature on fair clustering algorithms.
Chierichetti \emph{et al.} \cite{Chierichetti0LV17} first introduced the notion of \emph{proportional fairness} for clustering and then proposed fairlet decomposition algorithms for fair $k$-median and $k$-center clustering in the case of two groups.
Following this work, the problem has been further generalized to handle more than two groups \cite{HuangJV19}, permit lower and upper bounds on the fraction of points from a group \cite{Bercea0KKRS019, BeraCFN19, Gupta_2023}, and allow probabilistic group memberships \cite{EsmaeiliBT020}.
In addition, more efficient fair clustering algorithms have been proposed based on faster fairlet computation \cite{BackursIOSVW19} or coresets \cite{HuangJV19, SchmidtSS19}.
Then, several studies \cite{DavidsonR20, ZikoYGA21, EsmaeiliBSD21} explored how to achieve better trade-offs between the fairness and clustering objectives.
Other fair variants of clustering problems focused on different fairness notions, including individual-level fairness \cite{MahabadiV20, NegahbaniC21, VakilianY22}, fair center selection \cite{KleindessnerAM19, ThejaswiOG21}, and minimax losses among groups \cite{MakarychevV21, GhadiriSV21, ChlamtacMV22}.
However, all the above methods are primarily designed for i.i.d.~data rather than graph data.
They cannot be directly applied to graph partitioning because graphs are high-dimensional and sparse, and they will incur huge computational costs and often return inferior results due to the curse of dimensionality.

There were also a few studies on fairness in spectral methods and other graph clustering problems.
In addition to \cite{KleindessnerSAM19, pmlr-v206-wang23h}, Gupta and Dukkipati \cite{gupta2022consistency} further investigated spectral clustering with individual fairness constraints.
Moreover, Ahmadian \emph{et al.} \cite{AhmadianE0M20}, Friggstad and Mousavi \cite{FriggstadM21a}, and Ahmadian and Negahbani \cite{abs-2206-05050} studied fair correlation clustering on signed graphs.
Anagnostopoulos \emph{et al.} \cite{Anagnostopoulos20} applied spectral methods to fair densest subgraph discovery on graphs.
These algorithms are interesting but not comparable to our algorithm.

\section{Problem Definition}
\label{sec-def}

\smallskip\noindent\textbf{Normalized-Cut Graph Partitioning.}
Define $[n] := \{1, 2,$ $\ldots, n\}$ for any positive integer $n$.
Let $G = (V, E)$ be an undirected graph, where $V = [n]$ is the set of $n$ nodes and $E \subseteq V \times V$ is the set of edges.
We use $ \bm{W} = (w_{ij})_{i,j \in [n]}$ to denote the adjacency matrix of $G$, where each entry $w_{ij} \geq 0$ is the weight of edge $(i,j)$ (always equal to $1$ for the unweighted case) if it exists or $0$ otherwise.
We consider $w_{ii} = 0$ for any $i \in [n]$.
The degree matrix $\bm{D} = (d_{i})_{i \in [n]}$ is a diagonal matrix with the degree $d_{i} = \sum_{j=1}^n w_{ij} \geq 0$ of each node $i$ on its diagonal.
Given an undirected graph $G$, an integer $k \geq 2$, we aim to find a partitioning $\mathcal{C} = \{C_1, \ldots, C_k\}$ of $V$ into $k$ disjoint clusters, i.e., $\bigcup_{l = 1}^{k} C_l = V$ and $C_l \cap C_{l'} = \emptyset$ for any $l \neq l' \in [k]$, to minimize the normalized cut \cite{YuS03} (Ncut) value as follows:
\begin{equation}\label{eq-Ncut}
  \mathsf{Ncut}(\mathcal{C}) := \sum_{l=1}^k \frac{\mathsf{cut}(C_l)}{\mathsf{vol}(C_l)} = \sum_{l=1}^k \frac{\sum_{i \in C_l, j \in V \setminus C_l} w_{ij}}{\sum_{i \in C_l, j \in V} w_{ij}}.
\end{equation}
The Ncut minimization problem is well-known as NP-hard \cite{ShiM00}.

\smallskip\noindent\textbf{Fairness Constraint.}
In the fair variant of graph partitioning, we consider that the node set $V$ consists of several demographic groups defined by a categorical sensitive attribute, e.g., gender or race.
Formally, suppose that $V$ is divided into $m$ disjoint groups indexed by $[m]$, and an indicator function $\phi: [n] \mapsto [m]$ maps each node $i \in [n]$ to the group $\phi(i)$ it belongs to.
Let $ V_c = \{ i \in [n] : \phi(i) = c \} $ be the subset of nodes from group $c$ in $V$.
We assume that $ \bigcup_{c = 1}^{m} V_c = V $ and $ V_c \cap V_{c'} = \emptyset, \forall c \neq c' \in [m]$.
For ease of presentation, we denote the group membership as an indicator matrix $\bm{M} \in \{0,1\}^{n \times m}$, where $\bm{M}_{i,c} = 1$ if $\phi(i) = c $ and $0$ otherwise.
We follow a notion of \emph{range-based proportional fairness} in \cite{BeraCFN19} to require that every demographic group is approximately proportionally represented in all the $k$ clusters.
We define the fairness constraint by two vectors $\bm{\alpha}, \bm{\beta} \in [0,1]^{m}$ that specify the upper and lower bounds $\alpha_c, \beta_c$ on the percentage of nodes from group $c$.
We say a partitioning $\mathcal{C}$ is $(\bm{\alpha}, \bm{\beta})$-\emph{proportionally fair} if $\beta_c \leq \frac{\lvert V_c \cap C_l \rvert}{\lvert C_l \rvert} \leq \alpha_c$ for any $C_l$ and $V_c$.
In practice, we parameterize $\bm{\alpha}, \bm{\beta}$ by a fairness variable $\sigma \in [0, 1]$ as $\alpha_c = \min\{r_c / (1 - \sigma), 1\}$ and $\beta_c = r_c \cdot (1 - \sigma)$, where $r_c = |V_c| / n$.\footnote{Any other parameterization scheme (e.g., $\beta_c = r_c \cdot (1 - \sigma)$ and $\alpha_c = \beta_c + \sigma$) is also compatible with our formulation as long as it guarantees that $\beta_c \leq \alpha_c$ and $\beta_c, \alpha_c \in [0, 1]$ for any $c \in [m]$.}
For example, if the percentage of females is 60\% in the population of all nodes, the fairness constraint requires that the percentage of females in each cluster should be between 48\% and 75\% when $\sigma = 0.2$.
The value of $\sigma$ can be interpreted as how loose the fairness constraint is, where $\sigma = 0$ corresponds to every group in each cluster having the same ratio as that group in the population, and $\sigma = 1$ corresponds to no fairness constraint at all.
Given all the above notions, we formally define the normalized-cut graph partitioning problem under $(\bm{\alpha}, \bm{\beta})$-proportional fairness as follows.
\begin{definition}\label{def-fnm}
  Given an undirected graph $G = (V, E)$, a set of $m$ groups $V_1, \ldots, V_m \subseteq V$, two fairness vectors $ \bm{\alpha}, \bm{\beta} \in [0,1]^m$, and an integer $k \geq 2$, find an $(\bm{\alpha}, \bm{\beta})$-proportionally fair partitioning $ \mathcal{C} = \{C_1, \ldots, C_k \} $ of $V$ into $k$ disjoint clusters such that $\mathsf{Ncut}(\mathcal{C})$ in Eq.~\ref{eq-Ncut} is minimized.
\end{definition}
The problem in Definition~\ref{def-fnm} is NP-hard since the vanilla Ncut minimization problem is its special case when $m = 1$.
Next, we will approach the problem by extending the spectral Ncut minimization algorithm.
Note that our method can be adapted to \emph{ratio cut} \cite{HagenK92} and any other cut measure with an equivalent spectral formulation.
This paper focuses on Ncut due to its prevalence and space limitations.

\section{Our Algorithm}

We now introduce our two-phase spectral algorithm, FNM, for the problem of normalized-cut minimization with range-based proportional fairness constraints.
Next, we will present our range-based fair spectral embedding and rounding methods in Sections~\ref{subsec-spectral} and~\ref{subsec-rounding}.

\subsection{Range-based Fair Spectral Embedding}
\label{subsec-spectral}

\noindent\textbf{(Unconstrained) Spectral Normalized-Cut Minimization.}
We begin with a review of (unconstrained) spectral normalized-cut minimization.
According to \cite{YuS03}, the Ncut value can be expressed in terms of the graph Laplacian $\bm{L}:= \bm{D} - \bm{W}$ and a cluster membership indicator matrix $\bm{H} \in \mathbb{R}^{n \times k}$ as $ \mathsf{Ncut}(\mathcal{C}) = \mathsf{trace}(\bm{H}^\top \bm{L} \bm{H})$, where
\begin{equation}\label{H_matrix}
  \bm{H}_{i,l} =
  \begin{cases}
    \frac{1}{\sqrt{\mathsf{vol}(C_l)}}, & \text{if } \ i \in C_l; \\
    0, & \text{otherwise};
  \end{cases}
  \quad
  \forall i \in [n],\forall l \in [k].
\end{equation}
As such, the Ncut minimization problem is equivalent to minimizing $\mathsf{trace}(\bm{H}^\top \bm{L} \bm{H})$ over all possible $\bm{H}$ in the form of Eq.~\ref{H_matrix}.
However, the transformed problem is still NP-hard due to its combinatorial nature.
Therefore, the (normalized) spectral method \cite{YuS03} solves the following relaxed continuous optimization problem by allowing fractional assignments of nodes to clusters:
\begin{equation}\label{NM}
  \min_{\bm{H} \in \mathbb{R}^{n \times k}} \mathsf{trace}(\bm{H}^\top \bm{L} \bm{H})\;\;\text{s.t.}\;\;\bm{H}^\top \bm{D} \bm{H} = \bm{I}_k,
\end{equation}
where $\bm{I}_k$ is an identity matrix of size $k \times k$.
Note that $\bm{H}$ in the form of Eq.~\ref{H_matrix} must satisfy $\bm{H}^\top \bm{D} \bm{H} = \bm{I}_k$.
To solve the problem in Eq.~\ref{NM}, under an assumption that $d_{i}>0, \forall i \in [n]$ (i.e., $G$ has no isolated node), we substitute $\bm{H}$ with $\bm{D}^{-\frac{1}{2}}\bm{T}$ as follows:
\begin{equation}\label{NM1}
  \min_{\bm{T} \in \mathbb{R}^{n \times k}} \mathsf{trace}(\bm{T}^\top \bm{D}^{-\frac{1}{2}} \bm{L} \bm{D}^{-\frac{1}{2}} \bm{T}) \;\;\text{s.t.}\;\; \bm{T}^\top \bm{T} = \bm{I}_k.
\end{equation}
By Rayleigh-Ritz theorem \cite[\S~5.2.2]{Lutkepohl97}, an optimal solution to the problem in Eq.~\ref{NM1} is the matrix $\bm{T}$ which has the eigenvectors of $\bm{D}^{-\frac{1}{2}} \bm{L} \bm{D}^{-\frac{1}{2}}$ 
with respect to its $k$ smallest eigenvalues as columns.

\smallskip\noindent\textbf{Range-based Fair Spectral Ncut Minimization.}
We incorporate range-based proportional fairness into the problem in Eq.~\ref{NM1}.
Let us define two matrices $\bm{A} = [\bm{\alpha}, \ldots, \bm{\alpha}]^\top, \bm{B} = [\bm{\beta}, \ldots, \bm{\beta}]^\top \in \mathbb{R}^{n \times m}$ with the two fairness vectors $\bm{\alpha}, \bm{\beta} \in [0,1]^{m}$ as their rows.
By definition, a partitioning $ \mathcal{C} = \{C_1, \ldots, C_k\}$ denoted as $\bm{H}$ in the form of Eq.~\ref{H_matrix} is fair if and only if $(\bm{A} - \bm{M})^\top \bm{H} \geq \bm{0}$ and $(\bm{M} - \bm{B})^\top \bm{H} \geq \bm{0}$, where $\bm{0}$ is the zero matrix of size $m \times k$, because $\beta_c \cdot \lvert C_l \rvert \leq \lvert V_c \cap C_l \rvert \leq \alpha_c \cdot \lvert C_l \rvert$ if and only if the $(c,j)$-th entries of $(\bm{M} - \bm{B})^\top \bm{H}$ and $(\bm{A} - \bm{M})^\top \bm{H}$ are nonnegative for any $c \in [m]$ and $l \in [k]$.
In this way, the fairness constraints are expressed equivalently as \emph{linear constraints} in matrix form.

Given the above result, the problem in Definition~\ref{def-fnm} is transformed to minimizing $\mathsf{trace}(\bm{H}^\top \bm{L} \bm{H})$ over all $\bm{H}$ in the form of Eq.~\ref{H_matrix} with two additional linear constraints of $(\bm{A} - \bm{M})^\top \bm{H} \geq \bm{0}$ and $(\bm{M} - \bm{B})^\top \bm{H} \geq \bm{0}$.
By applying the same relaxation procedure as for the unconstrained problem, we obtain the following relaxed problem for range-based fair spectral Ncut minimization:
\begin{subequations}
\label{FNM}
\begin{align}
  \min_{\bm{T} \in \mathbb{R}^{n \times k}} & \quad \mathsf{trace}( \bm{T}^\top \bm{D}^{-\frac{1}{2}} \bm{L} \bm{D}^{-\frac{1}{2}} \bm{T}) \label{contiNM} \\
  \text{subject to} & \quad \bm{T}^\top \bm{T} = \bm{I}_k \label{Stiefel} \\
  & \quad (\bm{A} - \bm{M})^\top \bm{D}^{-\frac{1}{2}} \bm{T} \geq \bm{0} \label{alpha} \\
  & \quad (\bm{M} - \bm{B})^\top \bm{D}^{-\frac{1}{2}} \bm{T} \geq \bm{0} \label{beta}
\end{align}
\end{subequations}
Unlike Eq.~\ref{NM1}, the problem in Eq.~\ref{FNM} cannot be directly solved by eigendecomposition due to two additional constraints in Eqs.~\ref{alpha} and~\ref{beta}.

\smallskip\noindent\textbf{Range-based Fair Spectral Embedding with Augmented Lagrangian Method and OptStiefelGBB.}
To resolve the problem in Eq.~\ref{FNM}, we propose a novel algorithm based on the augmented Lagrangian method \cite{Wright99} for constrained optimization and OptStiefelGBB \cite{WenY13} for optimization with orthogonal constraints, as presented in Algorithm~\ref{alg1}, to find a solution $\bm{T}$ and a matrix $\bm{H} = \bm{D}^{-\frac{1}{2}} \bm{T}$ denoting a fractional assignment of each node in $V$ to $k$ clusters.

\begin{algorithm}[ht]
  \caption{Range-based Fair Spectral Embedding}\label{alg1}
  \textbf{Input}: Graph $G$ with adjacency matrix $\bm{W} \in \mathbb{R}^{n \times n}$, group matrix $\bm{M} \in \mathbb{R}^{n \times m}$, fairness vectors $\bm{\alpha}, \bm{\beta} \in [0,1]^m$, an integer $k \geq 2$\\
  \textbf{Parameters}: $T_1$, $\bm{\Lambda}_0$, $\mu_0$, $\xi$, $\varepsilon_1$; $T_2$, $\tau$, $\varepsilon_2$\\
  \textbf{Output}: Embedding matrix $\bm{H} \in \mathbb{R}^{n \times k}$
  \begin{algorithmic}[1]
    \STATE Compute $\bm{D}, \bm{L}$ and $\bm{A}, \bm{B}$ based on $\bm{W}$ and $\bm{\alpha}, \bm{\beta}$, respectively.
    \STATE Initialize an arbitrary matrix $\bm{T}_0 \in \mathbb{R}^{n \times k}$ with $\bm{T}_0^{\top} \bm{T}_0 = \bm{I}_k$.
    \FOR{$t = 0, 1, \ldots, T_1$}
    \STATE Formulate the problem in Eq.~\ref{subproblem} with $\bm{T}_{t}$, $\bm{\Lambda}_{t}$, and $\mu_{t}$.
    \REPEAT
      \STATE Use Eq.~\ref{Cayley} or~\ref{SMW} to compute $\bm{T}'_{t}$ w.r.t.~$\bm{T}_{t}$ and $\tau$.
      \STATE Set $\bm{T}_{t} \gets \bm{T}'_{t}$ and update $\tau$ using the Barzilai-Borwein method~\cite{Barzilai88}.
    \UNTIL{$\lVert \bm{\nabla}_{\bm{T}_t} L_{\mu_{t}}(\bm{T}_{t}, \bm{\Lambda}_{t}) \rVert_{F} \leq \varepsilon_2$ or after $T_2$ iterations}
    \STATE Set $\bm{T}_{t+1} \gets \bm{T}_{t}$ and update $\bm{\Lambda}_{t+1}, \mu_{t+1}$ based on Eq.~\ref{params}.
    \IF{$ \lVert \min \{\bm{P}(\bm{T}_{t+1}), \bm{0}\} \rVert_F \leq \varepsilon_1$}
      \STATE $\bm{H} \gets \bm{D}^{-\frac{1}{2}}\bm{T}_{t+1}$ and \textbf{break}.
    \ENDIF
    \ENDFOR
    \RETURN $\bm{H}$
  \end{algorithmic}
\end{algorithm}

The basic idea of the augmented Lagrangian method is to solve a constrained optimization problem by converting the constraints into penalty and Lagrange multiplier terms in the objective function.
In our problem, the violation of fairness constraints in Eqs.~\ref{alpha} and~\ref{beta} by a matrix $\bm{T}$ is denoted as the following penalty matrix:
\begin{displaymath}
  \bm{P}(\bm{T}) := [(\bm{A} - \bm{M})^\top \bm{D}^{-\frac{1}{2}}\bm{T}, (\bm{M} - \bm{B})^\top \bm{D}^{-\frac{1}{2}}\bm{T}] \in \mathbb{R}^{m \times 2k}.
\end{displaymath}
Based on \cite[\S~17.4]{Wright99}, the objective function in Eq.~\ref{contiNM} with penalty and Lagrange multiplier terms is as follows:
\begin{equation}\label{obj-Lagrange}
  L_{\mu}(\bm{T}, \bm{\Lambda}) := \mathsf{trace}( \bm{T}^\top \bm{D}^{-\frac{1}{2}} \bm{L} \bm{D}^{-\frac{1}{2}} \bm{T}) + \sum_{c=1}^m \sum_{l=1}^{2k} \rho(\bm{P}_{c,l}(\bm{T}), \bm{\Lambda}_{c,l}, \mu),
\end{equation}
where $\mu > 0$ is the penalty parameter, $\bm{\Lambda}$ is the estimation of Lagrange multipliers, and $\rho$ is defined as:
\begin{displaymath}
  \rho(p,\lambda,\mu) :=
  \begin{cases}
    -\lambda p + \frac{1}{2} \mu p^2, & \text{if}\; p - \frac{\lambda}{\mu} \leq 0; \\
    -\frac{1}{2\mu} \lambda^2, & \text{otherwise}.
\end{cases}
\end{displaymath}
The augmented Lagrangian method starts from the initial parameters $\mu_0 > 0$, $\bm{\Lambda}_0 = \bm{0}_{m \times 2k}$ and a solution $\bm{T}_0 \in \mathbb{R}^{n \times k}$ with $\bm{T}_0^{\top} \bm{T}_0 = \bm{I}_k$.
Then, it solves a sub-problem and iteratively updates the parameters.
Here, the $t$-th sub-problem ($t = 0, 1, \ldots$) to solve is:
\begin{equation}\label{subproblem}
  \min_{\bm{T} \in R^{n \times k}} L_{\mu_t}(\bm{T}, \bm{\Lambda}_t)\;\; \text{s.t.}\;\; \bm{T}^\top \bm{T} = \bm{I}_k.
\end{equation}
By solving Eq.~\ref{subproblem}, it obtains a new solution $\bm{T}_{t+1}$ and then updates $\mu$ and $\bm{\Lambda}$ as follows:
\begin{equation}\label{params}
    \bm{\Lambda}_{t+1} = \max\{\bm{\Lambda}_t - \mu_t \bm{P}(\bm{T}_{t+1}), \bm{0}\},
    \;\;
    \mu_{t+1} = \xi \mu_t,
\end{equation}
where $\xi > 1$ is an amplification parameter.
The augmented Lagrangian method terminates when the fairness violation of $\bm{T}_{t+1}$, defined as $\lVert \min \{\bm{P}(\bm{T}_{t+1}), \bm{0}\} \rVert_{F}$, is below an error parameter $\varepsilon_1 \geq 0$.
After that, we finally obtain a ``nearly fair'' embedding matrix $\bm{H} = \bm{D}^{-\frac{1}{2}} \bm{T}_{t+1}$ w.r.t.~$\bm{T}_{t+1}$.

Then, we consider how to solve the sub-problem in Eq.~\ref{subproblem} with OptStiefelGBB \cite{WenY13}, a general method for optimization under orthogonality constraints.
Its core idea is to model the feasible region as a $(n,k)$-Stiefel manifold \cite{manifold}, i.e., $\{\bm{T} \in \mathbb{R}^{n\times k}: \bm{T}^\top \bm{T} = \bm{I}_k\}$, and to apply Cayley transformation \cite{cayley} to update the solution at each iteration.
For the problem in Eq.~\ref{subproblem}, given a matrix $\bm{T} \in \mathbb{R}^{n \times k}$ with $\bm{T}^{\top} \bm{T} = \bm{I}_k$ and the gradients $\bm{\nabla_T} L_{\mu}(\bm{T}, \bm{\Lambda})$ ($\bm{\nabla_T}$ for short) of the augmented objective function $L_{\mu}(\bm{T}, \bm{\Lambda})$ w.r.t.~$\bm{T}$, the updated matrix $\bm{T}'$ is expressed as $\bm{T} - \frac{\tau}{2} \bm{\Omega} (\bm{T} + \bm{T}')$, where $\tau > 0$ is the step size and $\bm{\Omega} = \bm{\nabla_T} \bm{T}^{\top} - \bm{T} \bm{\nabla_T}^{\top}$.
By applying Cayley transformation \cite{cayley} on $\bm{\Omega}$, $\bm{T}'$ has a closed form expression as:
\begin{equation}\label{Cayley}
  \bm{T}' = (\bm{I}_n +\frac{\tau}{2} \bm{\Omega})^{-1}(\bm{I}_n - \frac{\tau}{2} \bm{\Omega}) \bm{T}.
\end{equation}
Then, we compute the gradients using the chain rule as follows:
\begin{equation}
\bm{\nabla_T} = 2\bm{D}^{-\frac{1}{2}} \bm{L} \bm{D}^{-\frac{1}{2}} \bm{T} + \bm{A}'{\bm{\nabla\rho}^{[0, k]}} + \bm{B}'{\bm{\nabla\rho}^{[k, 2k]}},
\end{equation}
where $\bm{A}' = ((\bm{A} - \bm{M})^\top \bm{D}^{-\frac{1}{2}})^\top$, $\bm{B}' = ((\bm{M} - \bm{B})^\top \bm{D}^{-\frac{1}{2}})^\top$, and
${\bm{\nabla\rho}^{[0, k]}}$, ${\bm{\nabla \rho}^{[k, 2k]}}$ are the first and last $k$ columns of $\bm{\nabla\rho} \in \mathbb{R}^{m \times 2k}$, respectively.
The $(c,l)$-entry of $\bm{\nabla\rho}$ is expressed as
\begin{equation}
  {(\bm{\nabla\rho})}_{c,l} =
  \begin{cases}
    -\bm{\Lambda}_{c,l} + \mu \bm{P}_{c,l}(\bm{T}), & \text{if } \bm{P}_{c,l} - \frac{\bm{\Lambda}_{c,l}}{\mu} \leq 0; \\
    0, & \text{otherwise.}
  \end{cases}
\end{equation}

Since computing the inversion of the matrix $\bm{I}_n +\frac{\tau}{2}\bm{\Omega} \in \mathbb{R}^{n \times n}$ is time-consuming, the Sherman-Morrison-Woodbury formula \cite[Appendix~A]{Wright99} is further applied to devise a much more efficient update scheme when $k \ll n$ that only computes the inversion of a much smaller matrix $\bm{I}_{2k} +\frac{\tau}{2} \bm{Y}^\top \bm{X} \in \mathbb{R}^{2k \times 2k}$ as follows:
\begin{equation}\label{SMW}
  \bm{T}' = \bm{T} - \tau \bm{X}(\bm{I}_{2k} + \frac{\tau}{2} \bm{Y}^\top \bm{X})^{-1} \bm{Y}^{\top} \bm{T}
\end{equation}
where $\bm{X} = [\bm{\nabla_T}, \bm{T}] \in \mathbb{R}^{n \times 2k}$ and $\bm{Y} = [\bm{T}, -\bm{\nabla_T}] \in \mathbb{R}^{n \times 2k}$.
According to \cite{WenY13}, $\bm{T}'$ by Eqs.~\ref{Cayley} and~\ref{SMW} has two important properties:
(\emph{i}) $\bm{T}'^{\top} \bm{T}' = \bm{I}_k$ if $\bm{T}^{\top} \bm{T} = \bm{I}_k$;
(\emph{ii}) $L_{\mu}(\bm{T}', \bm{\Lambda}) \leq L_{\mu}(\bm{T}, \bm{\Lambda})$.
Thus, with a proper step size $\tau$, OptStiefelGBB always converges to a feasible stationary point after sufficient iterations.
Following \cite{WenY13}, we use the Barzilai-Borwein method \cite{Barzilai88} to adaptively adjust the step size $\tau$ at each iteration.

\smallskip\noindent\textbf{Time Complexity.}
Let $T_1$ and $T_2$ denote the maximum numbers of iterations in the augmented Lagrangian method and OptStiefelGBB, respectively.
Computing the gradients $\bm{\nabla_T}$ for $L_{\mu}(\bm{T}, \bm{\Lambda})$ w.r.t.~$\bm{T}$ takes $O\left((|E| + n m) k\right)$ time.
Updating $\bm{T}$ with Eq.~\ref{Cayley} or~\ref{SMW} needs $O(n^3)$ or $O(n k^2)$ time, respectively.
As $m, k \leq n$ and $|E| \leq n^2$, the time complexity of Algorithm~\ref{alg1} is $O(T_1 T_2 n^3)$.
When $k \ll n$, $m = O(k)$, and $|E| = O(n)$, the time complexity of Algorithm~\ref{alg1} is reduced to $O(T_1 T_2 n k^2)$.

\subsection{Range-based Fair Rounding}
\label{subsec-rounding}
    
Like vanilla spectral methods, the output $\bm{H}$ of Algorithm~\ref{alg1} is a $k$-dimensional node embedding matrix where the $i$-th row vector $\bm{h}_i$ represents a fractional assignment of node $i \in [n]$ to $k$ clusters.
Thus, we must round the fractional solution into an integral one for partitioning.
However, a $k$-means clustering on embedding vectors, the standard rounding technique for spectral methods, is infeasible for the fair variant because the produced clusters may not be fair.

Next, we propose a novel rounding algorithm to produce a strictly fair partitioning scheme in Algorithm~\ref{alg2}.
Generally, it follows the same procedure as Lloyd's $k$-means clustering algorithm \cite{Lloyd82}, which initializes $k$ cluster centers, assigns each vector to one of the $k$ centers to generate the clusters, and updates each center to the median of each generated cluster iteratively until the stopping condition is met.
The difference from Lloyd's algorithm is that it requires the generated clusters at every iteration to be $(\bm{\alpha}, \bm{\beta})$-proportionally fair.

\begin{algorithm}[ht]
  \caption{Range-based Fair Rounding}\label{alg2}
  \textbf{Input}: Embedded vectors $H \subseteq \mathbb{R}^{k}$ (resp.~$V$), group indicator $\phi$ on $H$ and $V$, fairness vectors $\bm{\alpha}, \bm{\beta} \in [0,1]^m$, an integer $k \geq 2$\\
  \textbf{Parameters}: $T_3$, $\varepsilon_3$\\
  \textbf{Output}: Partitioning $\mathcal{C}^* = \{C^*_1, \ldots, C^*_k\}$
  \begin{algorithmic}[1]
  \STATE Initialize $\mathcal{C}^* = \emptyset$ and $t = 0$.
  \STATE Run $k$-means++ on $H$ to obtain an initial set $Q$ of centers.
  \REPEAT
    \STATE Set $t \gets t + 1$.
    \STATE Compute $\bm{C}$ w.r.t.~$H$, $Q$ and solve $\mathrm{LP1}$ to obtain $\bm{S}^*$.
    \STATE Assign each vector $\bm{h}_i \in H$ to a center $\varphi(\bm{h}_i) = \bm{q}_{l^*} \in Q$, where $l^* = \argmax_{l \in [k]} \bm{S}^{*}_{i, l}$.
    \STATE Solve $\mathrm{IP2}$ to obtain $\bm{N}'$ and compute $\bm{\Delta} = \bm{N}' - \bm{N}$.
    \WHILE{there is any $c \in [m]$ and $l \in [k]$ with $\Delta_{cl} \neq 0$}
    \STATE Pick arbitrary $c$, $C_l$, $C_{l'}$ with $\Delta_{cl} < 0$ and $\Delta_{cl'} > 0$.
    \STATE Let $\mathsf{cand} = \{ i \in [n]: \phi(i) = c \wedge \varphi(\bm{h}_i) = \bm{q}_l\}$.
    \STATE Set $i^* \gets \argmin_{i \in \mathsf{cand}} \delta_{l,l'}(i)$ and $\varphi(\bm{h}_{i^*}) = \bm{q}_{l'}$.
    \STATE Update $\bm{N}$ and $\bm{\Delta}$ for the reassignment of $i^*$.
    \ENDWHILE
    \STATE Generate a partitioning $\mathcal{C} = \{C_1, \ldots, C_k\}$ from $\varphi$, where $C_l = \{i \in [n] : \varphi(\bm{h}_i) = \bm{q}_l\}$ for each $l \in [k]$.
    \IF{$\mathcal{C}^* = \emptyset$ or $\mathsf{Ncut}(\mathcal{C}^*) > \mathsf{Ncut}(\mathcal{C})$}
    \STATE $\mathcal{C}^* \gets \mathcal{C}$.
    \ENDIF
    \STATE Set $Q' \gets Q$ and $Q \gets \{\bm{q}_l = \frac{1}{|C_l|} \sum_{i \in C_l} \bm{h}_i : l \in [k]\}$.
  \UNTIL{$\sum_{l = 1}^{k} \lVert \bm{q}_l - \bm{q}'_l \rVert \leq \varepsilon_3 $ for $\bm{q}_l \in Q, \bm{q}'_l \in Q'$ or $t \geq T_3$}
  \RETURN $\mathcal{C}^*$
  \end{algorithmic}
\end{algorithm}

\smallskip\noindent\textbf{Nearly-Fair Initial Assignment via LP.}
Our rounding algorithm begins with running $k$-means++ \cite{ArthurV07} on the set $H = \{ \bm{h}_i \in \mathbb{R}^k: i \in [n] \}$ of embedding vectors\footnote{The nodes in $V$ and vectors in $H$, as well as the groups defined on $V$ and $H$, will be used interchangeably in this subsection.} to obtain an initial set $Q = \{ \bm{q}_1, \ldots, \bm{q}_k \}$ of centers at the first iteration.
Then, we formulate the following \emph{fair assignment} problem \cite{BeraCFN19} to assign the vectors in $H$ to $Q$ to minimize the $l_2$-loss while ensuring that the cluster around each center is $(\bm{\alpha}, \bm{\beta})$-proportionally fair:
\begin{definition}\label{fair-assi}
  Given a point set $H$ with $m$ disjoint groups $H_1, \ldots,$ $H_m$, a set $Q$ of $k$ centers, and two fairness vectors $ \bm{\alpha}, \bm{\beta} \in [0,1]^m$, find an assignment $\varphi: H \to Q$ that minimizes $\sum_{\bm{h} \in H} \lVert \bm{h} - \varphi(\bm{h}) \rVert_2$ and ensures that $\beta_{c} \cdot \lvert C_{l} \rvert \leq \lvert H_c \cap C_l \rvert \leq \alpha_c \cdot \lvert C_l \rvert, \forall c \in [m], l \in [k]$, where $C_l = \{ \bm{h} \in H : \varphi(\bm{h}) = \bm{q}_l \}$.
\end{definition}

By denoting an assignment as an indicator matrix $\bm{S} \in \{0, 1\}^{n \times k}$, where $\bm{S}_{i,l} = 1$ if $\varphi(\bm{h}_i) = \bm{q}_l$ and $0$ otherwise, the problem in Definition~\ref{fair-assi} is represented as the following integer program ($\mathrm{IP1}$):
\begin{subequations}
\label{assi}
\begin{align}
  \mathrm{IP1} := \min \;\; & \mathsf{trace}(\bm{C}^{\top} \bm{S}) \label{assi-obj} \\
  \text{subject to} \;\; & (\bm{A} - \bm{M})^{\top} \bm{S} \geq \bm{0} \label{assi-alpha} \\
  & (\bm{M} - \bm{B})^{\top} \bm{S} \geq \bm{0} \label{assi-beta} \\
  & \bm{S} \bm{1}_k = \bm{1}_n,\, \bm{1}_{n}^{\top} \bm{S} \geq \bm{1}^{\top}_k \label{assi-not-empty} \\
  & \bm{S} \in \{0,1\}^{n \times k}, \label{int}
\end{align}
\end{subequations}
where Eq.~\ref{assi-obj} denotes the minimization of the $l_2$-loss by setting the cost matrix $\bm{C} \in \mathbb{R}^{n \times k}$ with $\bm{C}_{i,l} = \lVert \bm{h}_{i} - \bm{q}_{l} \rVert_2$,
Eqs.~\ref{assi-alpha} and~\ref{assi-beta} represent the fairness conditions,
and the two constraints in Eq.~\ref{assi-not-empty} mean that each vector must be assigned to exactly one center, and each center must be assigned with at least one vector.
Since $\mathrm{IP1}$ is NP-hard \cite{ShmoysT93}, we relax it to a linear program ($\mathrm{LP1}$) by substituting the condition of Eq.~\ref{int} with $\bm{S} \in [0,1]^{n \times k}$.
After obtaining the optimal solution $\bm{S}^{*}$ to $\mathrm{LP1}$ using any LP solver, we assign each vector $\bm{h}_i$ to the center $\varphi(\bm{h}_i) = \bm{q}_{l^*}$ with $l^* = \argmax_{l \in [k]} \bm{S}^{*}_{i, l}$.

\smallskip\noindent\textbf{Reassignments to Generate Strictly Fair Clusters.}
Although the above assignment scheme can produce fairer clusters than standard $k$-means without fairness constraints, it may still violate the fairness conditions after rounding a fractional solution by $\mathrm{LP1}$ to an integral one.
Therefore, we must reassign some vectors (nodes) to other centers to produce a strictly fair partitioning scheme.
To reduce the quality loss led by reassignments, we should (\emph{i}) move as few nodes as possible and (\emph{ii}) find the node leading to the smallest Ncut growth at each reassignment.
For (\emph{i}), we should seek a fair assignment $\varphi'$ closest to the current assignment $\varphi$.
Given an assignment $\varphi$, we use a matrix $\bm{N} = (n_{cl})_{c \in [m], l \in [k]} \in \mathbb{R}^{m \times k}$, where $n_{cl} = \lvert H_c \cap C_l \rvert$, to denote the number of nodes from each of the $m$ groups in $k$ clusters.
Then, we define the problem of computing an optimal scheme with the least number of reassigned nodes in the following integer program ($\mathrm{IP2}$):
\begin{subequations}
\begin{align}
  \mathrm{IP2} := \min \;\; & \sum_{c = 1}^m \sum_{l=1}^k \vert n'_{cl} - n_{cl} \vert \\
  \text{subject to} \;\; & \beta_c n'_{l} \leq n'_{cl} \leq \alpha_c n'_{l}, \forall c \in [m], l \in [k] \label{N-fair}\\
  & \sum_{l=1}^k n'_{cl} - n_{cl} = 0, \forall c \in [m] \\
  & n'_{cl} \in \mathbb{Z}^+, \forall c \in [m], l \in [k]
\end{align}
\end{subequations}
where $n'_{cl}$ is the number of nodes in $C_l$ from $H_c$ after reassignments, $n'_{l} = \sum_{c=1}^{m} n'_{cl}$, and the objective value is twice as many as the number of reassignments from $\varphi$ to fair $\varphi'$.
To solve $\mathrm{IP2}$, we can call an exact IP solver to find its optimal solution, which may take exponential time in the worst case but is still efficient in practice because the values of $k$ and $m$ are pretty small or run a heuristic search method, e.g., hill-climbing, to obtain a near-optimal solution in polynomial time.
Using either method, we can find a reassignment scheme denoted as $\bm{\Delta} = \bm{N}' - \bm{N}$, where $\Delta_{cl} = n'_{cl} - n_{cl}$ is greater than $0$ if $\Delta_{cl}$ nodes from $V_c$ should be moved to $C_l$, is smaller than $0$ if $-\Delta_{cl}$ nodes from $V_c$ should be moved from $C_l$, or is equal to $0$ if no reassignment is needed.
For (\emph{ii}), we first select a pair of clusters $C_{l}$ and $C_{l'}$ with $\Delta_{cl} < 0$ and $\Delta_{cl'} > 0$ for a specific $c$, which corresponds to the movement of a node in $V_c$ from $C_l$ to $C_{l'}$.
As reassignments can reduce partition quality, we want to find the node leading to as small Ncut growth as possible.
Since computing the Ncut value of a given partitioning from scratch is time-consuming and we only need to recalculate the changed parts (i.e., one node and two clusters) for a reassignment, we obtain the following equation to update the Ncut value incrementally by taking the difference between the Ncut values after and before reassigning a node $i$ from $C_l$ to $C_{l'}$ and eliminating all the unchanged terms:
\begin{equation}\label{eq-delta}
  \delta_{l,l'}(i) = \frac{\mathsf{cut}(C_l) - d_i + 2 z_{il}}{\mathsf{vol}(C_l) - d_i} - \frac{\mathsf{cut}(C_l)}{\mathsf{vol}(C_l)} + \frac{\mathsf{cut}(C_{l'}) + d_i - 2 z_{il'}}{\mathsf{vol}(C_{l'}) + d_i} - \frac{\mathsf{cut}(C_{l'})}{\mathsf{vol}(C_{l'})},
\end{equation}
where $z_{il} = \sum_{j \in C_l} w_{ij}$.
For each reassignment, we compute $\delta_{l,l'}(i)$ based on Eq.~\ref{eq-delta} for each eligible node $i$ (i.e., $i \in V_c \cap C_l$) and pick the node with the smallest $\delta_{l,l'}(i)$ accordingly.
After that, we update $\bm{N}, \bm{\Delta}$ and select the next group and pair of clusters for reassignment.
The above process terminates when $\bm{N} = \bm{N}'$ and all the clusters have been fair.
We obtain a fair partitioning $\mathcal{C} = \{C_1, \ldots, C_k\}$ from the final assignment, based on which we obtain an updated set $Q = \{\bm{q}_1, \ldots, \bm{q}_k\}$ of centers where $\bm{q}_l = \frac{1}{|C_l|} \sum_{i \in C_l} \bm{h}_i$.
After $Q$ is updated, the above procedures, i.e., solving $\mathrm{LP1}$ \& $\mathrm{IP2}$ and reassigning nodes for fairness, will be executed again to acquire a new fair partitioning.
This iterative procedure will terminate until the set $Q$ of centers does not change significantly between two iterations or the total number of iterations reaches a predefined threshold $T_3$.
Finally, a fair partitioning with the smallest Ncut value among all iterations will be returned as the final solution $\mathcal{C}^*$.

\smallskip\noindent\textbf{Time Complexity.}
The $k$-means++ algorithm takes $O(T_0 n k^2)$ time, where $T_0$ is the number of iterations.
At each iteration of Algorithm~\ref{alg2}, computing $\bm{C}$ takes $O(n k^2)$ time.
Using the interior point method \cite{Karmarkar84}, solving $\mathrm{LP1}$ takes $O(n^{4.5} k^{4.5} m)$ time in the worst case.
The hill-climbing search takes $O(n m k)$ time to solve $\mathrm{IP2}$.
Furthermore, the time to perform one reassignment is $O(n)$, and there are at most $O(n)$ reassignments.
But unlike Lloyd's $k$-means clustering algorithm, Algorithm~\ref{alg2} does not guarantee convergence to a local optimum after sufficient iterations.
If the convergence is not reached, it will stop and return the best partitioning found after $T_3$ iterations.
Suppose that $T_0 = O(T_3)$, the overall time complexity of Algorithm~\ref{alg2} is $O(T_3 n^{4.5} k^{4.5} m)$.

\section{Experiments}
\label{sec-exp}

In this section, we perform extensive empirical evaluations of our FNM algorithm.
We introduce our experimental setup in Section~\ref{setup} and describe our results in Section~\ref{results}.

\subsection{Experimental Setup}
\label{setup}

\noindent\textbf{Datasets.}
We use eight public real datasets with sensitive attributes and one synthetic dataset in the experiments.
\emph{Facebook}, \emph{LastFM}, \textit{Deezer}, \textit{Pokec-A}, \textit{Pokec-G} are all social networks;
\emph{DBLP} is a coauthor network;
\emph{German} and \emph{Credit} are similarity graphs created from i.i.d.~data;
and \emph{SBM} is generated from a stochastic block model with random groups.
If a graph is disconnected, we will extract and use its largest connected component.
Table~\ref{stats} summarizes the statistics of all processed datasets.
Detailed descriptions of the above datasets are provided in Appendix~\ref{app:dataset}.

\begin{table}[ht]
\footnotesize
\begin{center}
{\caption{Statistics of datasets in the experiments.}\label{stats}}
\begin{tabular}{|c|c|c|c|c|}
\hline
\textbf{Dataset} & $\lvert V \rvert$ & $\lvert E \rvert$ & \textbf{Sensitive Attribute} & $m$ \\
\hline
Facebook & 155 & 1,412 & gender & 2 \\
German & 1,000 & 21,742 & gender & 2 \\
SBM & 1,000 & 57,156 & -- & 5 \\
DBLP & 1,061 & 2,576 & continent & 3 \\
LastFM & 7,624 & 27,806 & country & 4 \\
Deezer & 28,281 & 92,752 & gender & 2 \\
Credit & 29,460 & 136,196 & education & 3 \\
Pokec-A & 1,097,077 & 10,792,894 & age & 4 \\
Pokec-G & 1,632,803 & 22,301,964 & gender & 2 \\
\hline
\end{tabular}
\end{center}
\end{table}

\smallskip\noindent\textbf{Baselines.}
We compare our FNM algorithm with the following three baseline methods for graph partitioning:
\emph{(i)} spectral clustering (SC) \cite{YuS03}, \emph{(ii)} fair spectral clustering (FSC) \cite{KleindessnerSAM19}, and \emph{(iii)} scalable fair spectral clustering (sFSC) \cite{pmlr-v206-wang23h}.
In the ablation study, we compare our range-based fair spectral embedding (rFSE) in Algorithm~\ref{alg1} with the following six node embeddings, i.e.,
spectral embedding (SE) \cite{YuS03}, fair spectral embedding (FSE) \cite{KleindessnerSAM19}, scalable fair spectral embedding (sFSE) \cite{pmlr-v206-wang23h}, DeepWalk (DW) \cite{PerozziAS14}, Node2Vec (N2V) \cite{GroverL16}, and FairWalk (FW) \cite{RahmanSBZ19};
and our range-based fair rounding (FR) in Algorithm~\ref{alg2} with four alternatives, i.e., $k$-means++ (KM+) \cite{ArthurV07}, $k$-means++ with reassignments (K+R), fair $k$-means (FK) \cite{BeraCFN19}, and solving $\mathrm{IP1}$ directly (IP).
Since the IP/LP solver fails to provide solutions in a reasonable time, FR, FK, and IP do not work on \textit{Pokec-A} and \textit{Pokec-G}, and we alternatively use K+R together with rFSE to obtain the results of FNM on both datasets.

\smallskip\noindent\textbf{Parameter Settings.}
For FNM,  $\bm{\alpha}, \bm{\beta}$ are parameterized by $\sigma \in [0,1]$ as per Section~\ref{sec-def}.
By default, we set $\sigma = 0.2$ and $0.8$ (resp.~the common 80\%-rule) to generate tight and loose fairness constraints.
For Algorithm~\ref{alg1}, we set $T_1 = 100$, $\varepsilon_1 = 10^{-6}$, $T_2 = 2,000$, $\tau = 10^{-3}$, and $\varepsilon_2 = 10^{-3}$.
We perform a grid search on $\xi \in \{2, 4, \ldots, 10\}$ and $\mu_0 \in \{10^{-4}, 10^{-2}, 10^0, 10^2\}$ and select the combination of $\xi, \mu_0$ achieving the lowest objective value for each experiment.
For Algorithm~\ref{alg2}, we set $T_0 = 100$, $T_3 = 10$, and $\varepsilon_3 = 10^{-4}$ for all experiments.
Further details of our parameter-tuning procedure are provided in Appendix~\ref{app:parameter}.
For the baselines, we follow the default parameters or use the recommended methods for parameter tuning as given in their original papers.

\smallskip\noindent\textbf{Evaluation Metrics.}
Each method is evaluated in three aspects.
First, we measure partition quality by the Ncut value in Eq.~\ref{eq-Ncut}.
Second, we adopt the notion of \emph{balance} in \cite{Chierichetti0LV17, BeraCFN19} as the metric for fairness.
Given a set $\mathcal{C} = \{C_1, \ldots, C_k\}$ of $k$ clusters and a set $\{V_1, \ldots, V_m\}$ of $m$ groups, the proportion of group $c$ in cluster $C_l$ is defined as $r_{cl} = \lvert C_l \cap V_c \rvert / \lvert C_l \rvert$.
Then, the \emph{balance} of $\mathcal{C}$ is defined by $\mathsf{balance}(\mathcal{C}) := \min_{c \in [m], l \in [k]} \min \big\{ r_c/r_{cl}, r_{cl}/r_c \big\}$, where $r_c = |V_c| / n$.
Higher \emph{balance} implies that the partitioning scheme is closer to being proportionally fair.
Balance also serves as an indicator of whether the fairness constraints parameterized by $\sigma$ are satisfied because $\mathcal{C}$ is $(\bm{\alpha}, \bm{\beta})$-proportionally fair iff $\mathsf{balance}(\mathcal{C}) \geq 1 - \sigma$.
Third, we use \emph{CPU time} to evaluate the efficiency of each method.

\smallskip\noindent\textbf{Implementation.}
We implement FNM in Python 3 and use Gurobi Optimizer to solve LP and IPs.
For each baseline, we either use a standard implementation in the SciPy library or the implementation published by the original authors.
The experiments were conducted on a desktop with an Intel Core i5-9500 processor @3.0GHz and 32GB RAM running Ubuntu 20.04.
Our code and data are published at \url{https://github.com/JiaLi2000/FNM}.

\subsection{Experimental Results}
\label{results}

\noindent\textbf{Overview.}
Table~\ref{default} presents the performance of different algorithms for normalized-cut graph partitioning with two fairness constraints parameterized by $\sigma = 0.8$ and $0.2$ when $k = 5$ on all nine datasets.

In terms of partition quality and fairness, the (unconstrained) SC mostly achieves the lowest Ncut values but fails to provide a fair partitioning when $\sigma = 0.8$ on four datasets while never meeting tighter fairness constraints when $\sigma = 0.2$.
Although FSC and sFSC provide more balanced partitions than SC in some cases, they still cannot guarantee the satisfaction of fairness constraints.
In addition, FSC does not return any results on medium and large graphs with over $10$k nodes due to huge memory consumption for eigendecomposition on dense matrices.
Next, we observe that FNM always provides fair partitioning schemes in all cases.
If unconstrained SC returns fair solutions when $\sigma = 0.8$, FNM will achieve nearly the same Ncut values.
Otherwise, the Ncut values of FNM will increase slightly to ensure fairness.
Moreover, the Ncut values of FNM for $\sigma = 0.2$ are significantly higher than those for $\sigma = 0.8$, which can be regarded as the \emph{price of fairness}.

In terms of time efficiency, FNM runs slower than SC as it is more time-consuming in embedding and rounding.
But FNM runs faster than FSC in most cases since it does not require eigendecomposition.
Compared to sFSC, which improves the scalability of FSC by avoiding eigendecomposition on dense matrices, FNM runs slower on smaller graphs due to a longer time for fair rounding but becomes faster on larger graphs owing to the efficiency improvements for fair embedding.
Finally, FNM is more efficient when $\sigma = 0.8$ than $\sigma = 0.2$ because of fewer iterations for convergence.

\begin{table}[t]
  \footnotesize
  \setlength\tabcolsep{1pt}
  \begin{center}
  {\caption{Performance of different algorithms for normalized-cut graph partitioning with $k = 5$ clusters. Cells in lighter and darker gray colors denote results satisfying looser ($\sigma = 0.8$) and tighter ($\sigma = 0.2$) fairness constraints, respectively. For the Ncut values, we highlight the best overall results on each dataset in \textbf{bold} font and \underline{underline} the best fair result when $\sigma = 0.8$. FSC is marked by ``--'' when it does not provide any solution due to huge memory consumption for eigendecomposition on dense matrices.}\label{default}}
  \begin{tabular}{|c|ccc|ccc|ccc|ccc|ccc|}
  \hline
  \multirow{2}{*}{\textbf{Dataset}} & \multicolumn{3}{c|}{\textbf{SC}} & \multicolumn{3}{c|}{\textbf{FSC}} & \multicolumn{3}{c|}{\textbf{sFSC}} & \multicolumn{3}{c|}{\textbf{FNM} ($\sigma = 0.8$)} & \multicolumn{3}{c|}{\textbf{FNM} ($\sigma = 0.2$)} \\
  \cline{2-16}
  & Ncut & Balance & Time (s) & Ncut & Balance & Time (s) & Ncut & Balance & Time (s) & Ncut & Balance & Time (s) & Ncut & Balance & Time (s) \\
  \hline
  Facebook & \cellcolor{gray!20}\textbf{1.378} & \cellcolor{gray!20}0.458 & \cellcolor{gray!20}0.233 & \cellcolor{gray!20}1.401 & \cellcolor{gray!20}0.623 & \cellcolor{gray!20}0.252 & \cellcolor{gray!20}1.401 & \cellcolor{gray!20}0.623 & \cellcolor{gray!20}0.193 & \cellcolor{gray!20}\textbf{1.378} & \cellcolor{gray!20}0.458 & \cellcolor{gray!20}0.225 & \cellcolor{gray!50}1.550 & \cellcolor{gray!50}0.81 & \cellcolor{gray!50}0.312 \\
  German   & \cellcolor{gray!20}\textbf{1.433} & \cellcolor{gray!20}0.211 & \cellcolor{gray!20}0.357 & \cellcolor{gray!20}1.442 & \cellcolor{gray!20}0.583 & \cellcolor{gray!20}0.817 & \cellcolor{gray!20}1.442 & \cellcolor{gray!20}0.583 & \cellcolor{gray!20}0.581 & \cellcolor{gray!20}\textbf{1.433} & \cellcolor{gray!20}0.211 & \cellcolor{gray!20}0.486 & \cellcolor{gray!50}1.498 & \cellcolor{gray!50}0.8 & \cellcolor{gray!50}2.841 \\
  SBM      & \cellcolor{gray!20}\textbf{2.542} & \cellcolor{gray!20}0.226 & \cellcolor{gray!20}0.375 & \cellcolor{gray!20}2.619 & \cellcolor{gray!20}0.245 & \cellcolor{gray!20}0.921 & \cellcolor{gray!20}2.619 & \cellcolor{gray!20}0.245 & \cellcolor{gray!20}0.703 & \cellcolor{gray!20}\textbf{2.542} & \cellcolor{gray!20}0.226 & \cellcolor{gray!20}0.660 & \cellcolor{gray!50}3.348 & \cellcolor{gray!50}0.81 & \cellcolor{gray!50}2.608 \\
  DBLP     & \textbf{0.022} & 0 & 0.317 & 0.024 & 0 & 0.568 & 0.024 & 0 & 0.435 & \cellcolor{gray!20}\underline{0.050} & \cellcolor{gray!20}0.2 & \cellcolor{gray!20}0.658 & \cellcolor{gray!50}0.269 & \cellcolor{gray!50}0.8 & \cellcolor{gray!50}0.698 \\
  LastFM   & \textbf{0.119} & 0 & 0.526 & 0.185 & 0 & 102.2 & 0.185 & 0 & 1.462 & \cellcolor{gray!20}\underline{0.265} & \cellcolor{gray!20}0.2 & \cellcolor{gray!20}4.624 & \cellcolor{gray!50}0.699 & \cellcolor{gray!50}0.8 & \cellcolor{gray!50}7.014 \\
  Deezer   & \cellcolor{gray!20}\textbf{0.038} & \cellcolor{gray!20}0.406 & \cellcolor{gray!20}2.989 & -- & -- & -- & \cellcolor{gray!20}0.040 & \cellcolor{gray!20}0.406 & \cellcolor{gray!20}8.246 & \cellcolor{gray!20}\textbf{0.038} & \cellcolor{gray!20}0.406 & \cellcolor{gray!20}6.465 & \cellcolor{gray!50}0.216 & \cellcolor{gray!50}0.81 & \cellcolor{gray!50}9.120 \\
  Credit   & \cellcolor{gray!20}0.035 & \cellcolor{gray!20}0.738 & \cellcolor{gray!20}9.097 & -- & -- & -- & \cellcolor{gray!20}0.035 & \cellcolor{gray!20}0.665 & \cellcolor{gray!20}41.28 & \cellcolor{gray!20}\textbf{0.034} & \cellcolor{gray!20}0.748 & \cellcolor{gray!20}10.74 & \cellcolor{gray!50}0.049 & \cellcolor{gray!50}0.8 & \cellcolor{gray!50}12.33 \\
  Pokec-A  & \textbf{0.077} & 0 & 241.5 & -- & -- & -- & \textbf{0.077} & 0 & 686.9 & \cellcolor{gray!20}\underline{0.450} & \cellcolor{gray!20}0.209 & \cellcolor{gray!20}270.6 & \cellcolor{gray!50}2.701 & \cellcolor{gray!50}0.8 & \cellcolor{gray!50}254.4 \\
  Pokec-G  & \textbf{0.070} & 0.141 & 445.8 & -- & -- & -- & \textbf{0.070} & 0.141 & 1024 & \cellcolor{gray!20}\underline{0.128} & \cellcolor{gray!20}0.254 & \cellcolor{gray!20}525.0 & \cellcolor{gray!50}1.330 & \cellcolor{gray!50}0.81 & \cellcolor{gray!50}644.8 \\
  \hline
  \end{tabular}
  \end{center}
\end{table}

\begin{figure}[t]
  \centering
  \subfloat{\includegraphics[width=\linewidth]{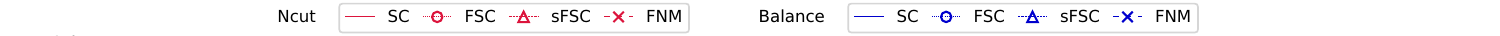}}
  \\
  \subfloat[Facebook]{\includegraphics[width=0.196\linewidth]{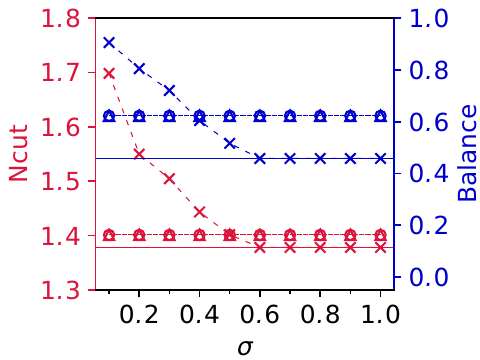}}
  \subfloat[SBM]{\includegraphics[width=0.196\linewidth]{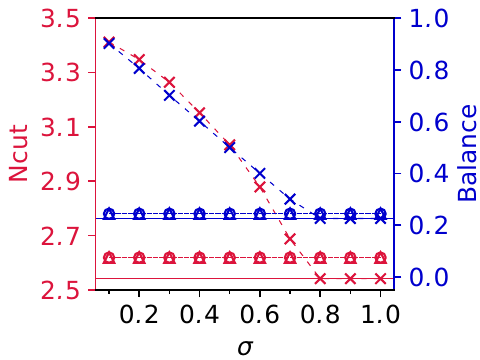}}
  \subfloat[DBLP]{\includegraphics[width=0.196\linewidth]{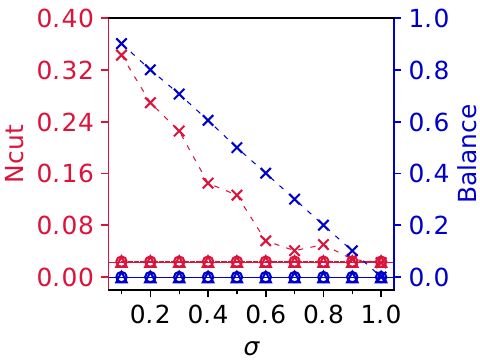}}
  \subfloat[LastFM]{\includegraphics[width=0.196\linewidth]{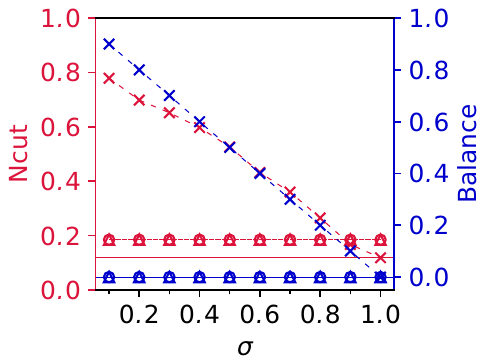}}
  \subfloat[Credit]{\includegraphics[width=0.196\linewidth]{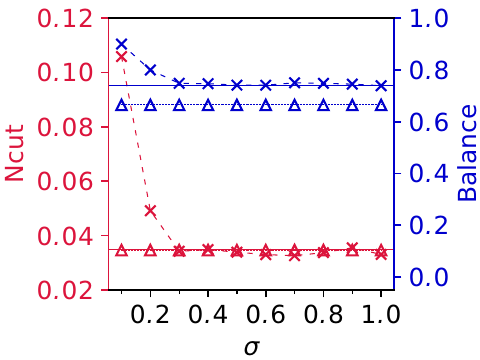}}
  \caption{Quality (i.e., Ncut) and fairness (i.e., balance) of $k=5$ partitions created by SC, FSC, sFSC, and FNM as a function of parameter $\sigma$. Note that when $\sigma$ varies, FNM offers trade-offs between fairness and quality but SC, FSC, and sFSC remain unchanged in both measures.}
  \label{fig-sigma}
\end{figure}

\smallskip\noindent\textbf{Trade-off between Quality and Fairness.}
We present the performance of four algorithms with different fairness constraints parameterized by $\sigma = 0.1, 0.2, \ldots, 1$ in Figure~\ref{fig-sigma}.
We ignore $\sigma = 0$ since no solution may exist for indivisibility.
Since the results of SC, FSC, and sFSC are independent of $\sigma$, they are drawn as horizontal lines in the figure.
For FNM, as the value of $\sigma$ decreases (when the fairness constraints become looser), the Ncut and balance values also decrease.
When $\sigma = 1$ (no fairness constraint), FNM returns partitions of similar quality to SC.
To our best knowledge, FNM is the only known algorithm that achieves different trade-offs between partition quality (i.e., \emph{Ncut}) and fairness (i.e., \emph{balance}) w.r.t.~$\sigma$.

\begin{figure}[t]
  \centering
  \subfloat{\includegraphics[width=\linewidth]{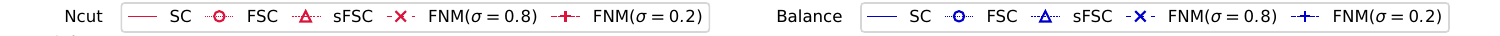}}
  \\
  \subfloat[Facebook]{\includegraphics[width=0.196\linewidth]{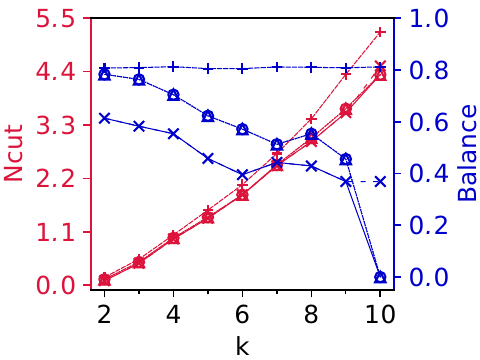}}
  \subfloat[SBM]{\includegraphics[width=0.196\linewidth]{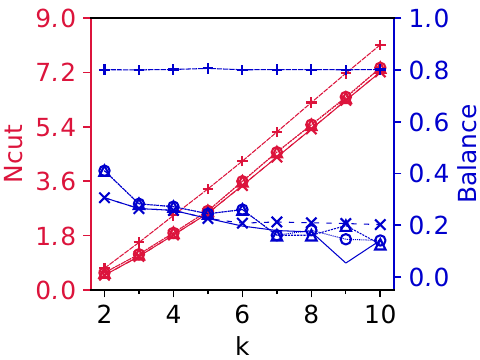}}
  \subfloat[DBLP]{\includegraphics[width=0.196\linewidth]{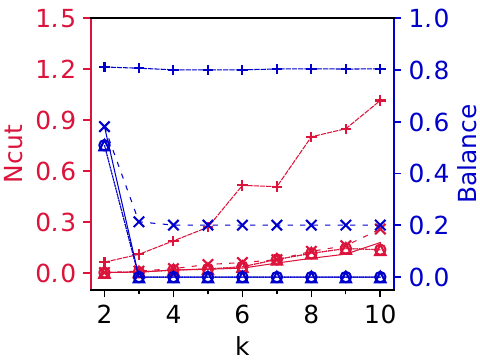}}
  \subfloat[LastFM]{\includegraphics[width=0.196\linewidth]{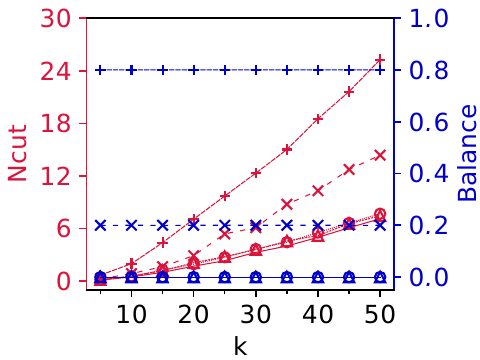}}
  \subfloat[Credit]{\includegraphics[width=0.196\linewidth]{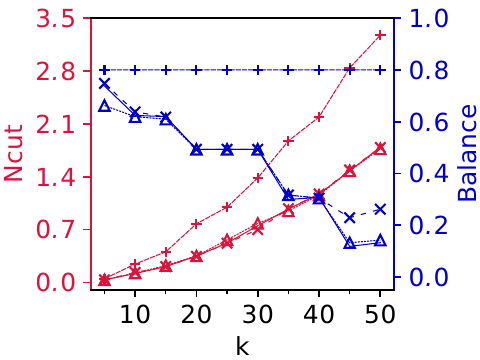}}
  \caption{Quality (i.e, Ncut) and fairness (i.e., balance) of SC, FSC, sFSC, and FNM ($\sigma=0.8, 0.2$) as a function of the number of clusters $k$.}
  \label{fig-k}
\end{figure}

We illustrate the performance of four algorithms as a function of the numbers of clusters $k$ in Figure~\ref{fig-k}.
We vary $k$ from $2$ to $10$ on three smaller datasets and from $5$ to $50$ on two larger datasets.
For each algorithm, the Ncut value increases with $k$.
Meanwhile, the balance of each algorithm generally drops with increasing $k$, and FNM is the only algorithm that consistently achieves a balance of at least $1 - \sigma$, i.e., guaranteeing the fairness constraints.
Furthermore, FNM has comparable Ncut values to SC, FSC, and sFSC and higher balances on most datasets when $\sigma = 0.8$.
But on the \emph{LastFM} dataset, FNM has higher Ncut values than other algorithms when $\sigma = 0.8$ because it performs reassignments to ensure a balance of at least $0.2$.
When $\sigma = 0.2$, as FNM assigns more nodes to non-closest clusters for fairness than when $\sigma = 0.8$, it often has inferior partition quality, especially when $k$ is large.

Note that the results examining the partition quality by varying $\sigma$ and $k$ on the remaining four datasets, as well as the time efficiency by varying $\sigma$ and $k$ on all the nine datasets, are deferred to Appendix~\ref{app:additional}.


\smallskip\noindent\textbf{Ablation Study.}
In the ablation study, we first run each embedding method to obtain a $k$-dimensional node embedding (for $k = 5, 20$) on each graph and use the same fair rounding in Algorithm~\ref{alg2} to produce two fair partitioning schemes with $\sigma = 0.8, 0.2$ from node vectors.
SC, FSC, sFSC, and FNM are thus renamed SE, FSE, sFSE, and rFSE since we only test their spectral embedding performance.
We report the Ncut values on five small datasets (since FSE and FW cannot provide any result on four large datasets) in Table~\ref{tbl-ablation-embed}.
Our method, rFSE in Algorithm~\ref{alg1}, achieves the best or second-best partition quality among all methods in almost all cases.
Here, SE and FSE/sFSE are special cases of rFSE when $\sigma = 1$ (i.e., w/o fairness) and $0$ (i.e., with strict proportionality), respectively.
As such, SE performs closely to rFSE when $\sigma = 0.8$, and FSE/sFSE provides similar embeddings to rFSE when $\sigma = 0.2$.
Especially, sFSE slightly outperforms rFSE when $\sigma = 0.2$. Since sFSE and rFSE both provide embeddings with fractional fairness constraints, which are looser than the original integral ones, using a tighter (fractional) constraint than required (i.e., equal to $1 - \sigma$) in the embedding phase might help improve the overall performance after rounding.
The partition quality of deep learning-based node embeddings (with or without fairness) is much inferior to that of rFSE and other spectral embeddings since they are not designed for graph partitioning.

Then, we evaluate the performance of each rounding method on node vectors provided by rFSE on five small datasets for $k = 5, 20$ and $\sigma = 0.8, 0.2$ and present the Ncut values in Table~\ref{tbl-ablation-round}.
Our method, FR in Algorithm~\ref{alg2}, performs best among the four fair methods we compare.
We find that fair $k$-means for i.i.d.~data in \cite{BeraCFN19} may not be appropriate to round embedding vectors though it adopts the same fairness notion as ours.
Despite having the lowest Ncut values, $k$-means++ cannot produce fair partitions.
When $\sigma = 0.8$, FR provides partitions closer to $k$-means++ than when $\sigma = 0.2$ since fewer or no reassignments are required as fairness constraints are looser.

\begin{table}[t]
\footnotesize
\setlength\tabcolsep{1pt}
\begin{center}
{\caption{Performance of different embedding methods on partition quality (i.e., \emph{Ncut values}). Here, the best result(s) are highlighted in a \textbf{bold} font, and the second best result(s) are \underline{underlined}.}\label{tbl-ablation-embed}}
\begin{tabular}{|c|c|c|c|c|c|c|c|c|c|c|c|c|c|c|c|}
    \hline
    \multirow{2}{*}{\textbf{Dataset}} &
    \multirow{2}{*}{$k$} &
    \multicolumn{7}{c|}{$\sigma=0.8$} &
    \multicolumn{7}{c|}{$\sigma=0.2$} \\ \cline{3-16} 
    & & \textbf{SE} & \textbf{FSE} & \textbf{sFSE} & \textbf{DW} & \textbf{N2V} & \textbf{FW} & \textbf{rFSE ($^*$)} & \textbf{SE} & \textbf{FSE} & \textbf{sFSE} & \textbf{DW} & \textbf{N2V} & \textbf{FW} & \textbf{rFSE ($^*$)} \\ \hline
    \multirow{2}{*}{Facebook} & 5
    & \underline{1.378} & 1.401 & 1.401 & \textbf{1.374} & 1.438 & 1.444 & \underline{1.378}
    & 1.676 & \textbf{1.536} & \textbf{1.536} & 1.676 & 1.575 & 1.689 & \underline{1.546} \\ \cline{2-16} 
    & 20
    & 13.840 & 13.887 & 14.057 & \textbf{13.050} & \underline{13.230} & 14.825 & 13.753
    & 14.861 & 14.882 & 14.882 & \underline{14.541} & 14.708 & 15.461 & \textbf{14.346} \\ \hline
    \multirow{2}{*}{German} & 5
    & \textbf{1.433} & 1.442 & 1.442 & 1.500 & 1.492 & 1.492 & \textbf{1.433}
    & 1.537 & \textbf{1.471} & \textbf{1.471} & 1.519 & 1.502 & 1.505 & \underline{1.498} \\ \cline{2-16} 
    & 20
    & \underline{11.856} & 11.879 & 11.869 & 11.977 & 11.922 & 12.638 & \textbf{11.811}
    & 12.927 & 12.889 & \underline{12.884} & 12.954 & 13.012 & 13.059 & \textbf{12.852} \\ \hline
    \multirow{2}{*}{SBM} & 5
    & \textbf{2.542} & 2.619 & 2.619 & 2.585 & 2.807 & 2.941 & \textbf{2.542}
    & 3.377 & \textbf{3.345} & \textbf{3.345} & 3.490 & 3.509 & 3.515 & \underline{3.348} \\ \cline{2-16} 
    & 20
    & \underline{16.812} & 16.998 & 17.012 & 17.294 & 17.357 & 17.430 & \textbf{16.785}
    & 17.844 & 17.812 & \textbf{17.784} & 18.131 & 18.194 & 18.104 & \underline{17.799} \\ \hline
    \multirow{2}{*}{DBLP} & 5
    & \underline{0.050} & \textbf{0.032} & \textbf{0.032} & 0.929 & 1.002 & 0.423 & \underline{0.050}
    & 1.003 & \textbf{0.261} & \textbf{0.261} & 0.995 & 1.029 & 0.645 & \underline{0.269} \\ \cline{2-16} 
    & 20
    & 1.381 & \textbf{0.941} & \underline{0.984} & 7.560 & 6.875 & 3.793 & 1.166
    & 3.170 & 2.871 & \textbf{2.779} & 8.691 & 8.119 & 6.235 & \underline{2.787} \\ \hline
    \multirow{2}{*}{LastFM} & 5
    & \underline{0.294} & 0.453 & 0.453 & 0.846 & 0.754 & 1.392 & \textbf{0.265}
    & 0.908 & \textbf{0.677} & \textbf{0.677} & 1.704 & 1.667 & 1.900 & \underline{0.699} \\ \cline{2-16} 
    & 20
    & \underline{3.983} & 4.666 & 4.943 & 7.313 & 6.942 & 8.316 & \textbf{2.922}
    & \underline{7.367} & 7.699 & 7.680 & 10.698 & 10.612 & 11.607 & \textbf{7.080} \\ \hline
    \multicolumn{2}{|c|}{Avg.~Ranking}
    & \underline{2.2} & 2.8 & 3.1 & 4.1 & 4.4 & 5.5 & \textbf{1.6}
    & 4.0 & 2.2 & \textbf{1.6} & 4.8 & 4.8 & 5.4 & \underline{1.7} \\ \hline
\end{tabular}
\end{center}
\end{table}

\begin{table}[t]
\footnotesize
\setlength\tabcolsep{1pt}
\begin{center}
{\caption{Performance of different rounding methods on partition quality (i.e., \emph{Ncut values}). Here, the best result(s) are highlighted in a \textbf{bold} font, and the second best result(s) are \underline{underlined}. Note that the results of $k$-means++ do not satisfy the fairness constraints and thus are just presented to show the ``price of fairness'' in the rounding process.}\label{tbl-ablation-round}}
\begin{tabular}{|c|c|c|c|c|c|c|c|c|c|c|c|}
    \hline
    \multirow{2}{*}{\textbf{Dataset}} &
    \multirow{2}{*}{$k$} &
    \multicolumn{5}{c|}{$\sigma=0.8$} &
    \multicolumn{5}{c|}{$\sigma=0.2$} \\ \cline{3-12} 
    & & $k$-means++ & \textbf{K+R} & \textbf{FK} & \textbf{IP} & \textbf{FR ($^*$)} & $k$-means++ & \textbf{K+R} & \textbf{FK} & \textbf{IP} & \textbf{FR ($^*$)} \\ \hline
    \multirow{2}{*}{Facebook} & 5 & (1.378) & \textbf{1.378} & \textbf{1.378} & \textbf{1.378} & \textbf{1.378} & (1.381) & 1.640 & \underline{1.576} & 1.599 & \textbf{1.546} \\ \cline{2-12} 
    & 20 & (12.998) & 14.161 & 13.947 & \textbf{13.333} & \underline{13.753} & (12.586) & 14.991 & 14.887 & \underline{14.615} & \textbf{14.346} \\ \hline
    \multirow{2}{*}{German} & 5 & (1.433) & \textbf{1.433} & 1.442 & \textbf{1.433} & \textbf{1.433} & (1.479) & 1.526 & \textbf{1.498} & 1.499 & \textbf{1.498} \\ \cline{2-12} 
    & 20 & (11.629) & 11.995 & 11.948 & \underline{11.883} & \textbf{11.811} & (11.733) & 13.611 & 13.166 & \underline{13.069} & \textbf{12.852} \\ \hline
    \multirow{2}{*}{SBM} & 5 & (2.542) & \textbf{2.542} & \textbf{2.542} & \textbf{2.542} & \textbf{2.542} & (2.568) & \underline{3.350} & \underline{3.350} & \underline{3.350} & \textbf{3.348} \\ \cline{2-12} 
    & 20 & (16.767) & 16.839 & 16.834 & \underline{16.803} & \textbf{16.785} & (16.908) & 18.006 & \underline{17.926} & 17.889 & \textbf{17.799} \\ \hline
    \multirow{2}{*}{DBLP} & 5 & (0.022) & 0.062 & 0.067 & \textbf{0.049} & \underline{0.050} & (0.024) & \underline{0.466} & 0.521 & 0.500 & \textbf{0.269} \\ \cline{2-12} 
    & 20 & (0.400) & 1.381 & \underline{1.066} & \textbf{1.007} & 1.166 & (0.404) & 4.596 & 4.104 & \underline{3.452} & \textbf{2.787} \\ \hline
    \multirow{2}{*}{LastFM} & 5 & (0.152) & 0.320 & 0.272 & \underline{0.266} & \textbf{0.265} & (0.184) & 0.801 & 0.805 & \underline{0.791} & \textbf{0.699} \\ \cline{2-12} 
    & 20 & (2.005) & 5.751 & 5.139 & \underline{3.211} & \textbf{2.922} & (2.257) & 10.723 & 11.144 & \underline{9.911} & \textbf{7.080} \\ \hline
    \multicolumn{2}{|c|}{Avg.~Ranking} & -- & 2.9 & 2.8 & \textbf{1.4} & \textbf{1.4} & -- & 3.4 & 2.8 & \underline{2.4} & \textbf{1.0} \\ \hline
\end{tabular}
\end{center}
\end{table}

\section{Conclusion}

This paper investigated the $(\bm{\alpha}, \bm{\beta})$-proportionally fair normalized cut graph partitioning problem.
We proposed a novel algorithm, FNM, consisting of an extended spectral embedding method and a $k$-means-based rounding scheme to provide a node partitioning with a small Ncut value on a graph while strictly following the proportional fairness constraints.
The comprehensive experimental findings confirmed the superior performance of FNM in terms of partition quality, fairness, and efficiency.
In future work, we will generalize our algorithm to handle other notions of fairness, e.g., individual fairness~\cite{MahabadiV20, gupta2022consistency}, in graph partitioning problems.

\section*{Ethics Statement}
The implication of this work is that it enables us to find proportionally representative graph partitions. This is relevant for various real-world applications to reduce harmful biases of traditional algorithms for this problem. As such, we do not foresee situations in which our method may be directly misused. This is based on the assumption that the user aims to improve social outcomes rather than having a negative impact.

\section*{Acknowledgements}
This work was supported by the National Natural Science Foundation of China under grant no.~62202169. We would like to thank the anonymous reviewers for their comments, which helped improve this work considerably.

\bibliography{refs}

\appendix

\section{Dataset Preparation}
\label{app:dataset}

Our experiments use eight public real-world datasets with sensitive attributes in \cite{10097603, snapnets} and one synthetic dataset.
Here, we provide detailed information on each dataset as follows.
\begin{itemize}
  \item \textbf{Facebook} \cite{Mastrandrea15} is a social network representing the friendships between students in a high school in Marseilles, France. We remove the nodes whose \textit{gender} attributes are unknown and extract the largest connected component with $155$ nodes and $1,412$ edges.
  \item \textbf{German}\footnote{\url{https://github.com/yushundong/Graph-Mining-Fairness-Data/tree/main/dataset/german}} is a similarity graph created from an i.i.d. dataset where each record contains both categorical and numerical attributes to predict a person's credit risk. It consists of $1,000$ nodes and $21,742$ edges with \textit{gender} as the sensitive attribute.
  \item \textbf{SBM} is a synthetic stochastic block model \cite{HOLLAND1983109} graph comprising $5$ clusters and $5$ groups with intra-cluster and inter-cluster probabilities of 0.25 and 0.05, respectively. For cluster sizes, we set $\vert C_1 \vert = 500$, $\vert C_2 \vert = 200$, $\vert C_3 \vert = 100$, $\vert C_4 \vert = 100$, $\vert C_5 \vert = 100$.
  For each node in cluster $C_l$, we assign it to group $l$ with probability $0.6$ or to any other group $l' \neq l$ with probability $0.1$.
  According to our generation procedure, the five ground-truth partitions all have balance values of around $0.2$. When $\sigma = 0.8$, the best partitioning scheme will be very close to the ground truth, and when $\sigma < 0.8$ (e.g., $0.2$), reassignments will be required to ensure fairness.
  Using the SBM graph, we aim to evaluate the performance of each algorithm in controllable settings. On the one hand, can it recover the ground-truth partitions when the fairness constraint is relatively loose? On the other hand, how much does it spend on modifying the ground-truth partitions to be fair (as measured by the Ncut increase) with relatively tighter constraints?
  \item \textbf{DBLP}\footnote{\url{https://github.com/yushundong/Graph-Mining-Fairness-Data/tree/main/dataset/dblp}} is a coauthor network, where there is an edge between two authors if they have ever collaborated. We merge the five original groups \{ `Asia', `Oceania', `North America', `South America', `Europe'\} by \emph{continent} into three new groups \{`Asia-Oceania', `America', `Europe'\}. We also extract the largest connected component with $1,061$ nodes and $2,576$ edges for evaluation.
  \item \textbf{LastFM}\footnote{\url{https://snap.stanford.edu/data/feather-lastfm-social.html}} is a social network with $7,624$ nodes and $27,806$ edges representing the friendships between users on the last.fm website. The nodes are divided into four groups by \emph{country}.
  \item \textbf{Deezer}\footnote{\url{https://snap.stanford.edu/data/feather-deezer-social.html}} is a social network with $28,281$ nodes and $92,752$ edges representing the friendships between users on Deezer, where \textit{gender} is used as the sensitive attribute.
  \item \textbf{Credit}\footnote{\url{https://github.com/yushundong/Graph-Mining-Fairness-Data/tree/main/dataset/credit}} is also a similarity graph created from an i.i.d. dataset for credit risk analysis. We extract the largest connected component with $29,460$ nodes and $136,196$ edges and use \emph{education} as the sensitive attribute.
  \item \textbf{Pokec}\footnote{\url{https://snap.stanford.edu/data/soc-Pokec.html}} is an online social network in Slovakia. The directed edges denoting the follower-followee relationships in the original graph are transformed into undirected edges. For sensitive attribute \textit{age}, we remove the nodes whose age information is not available and transform it into a category attribute as `1': [0, 18], `2': [19, 25], `3': [26,35], `4': 36+. Then, we obtain an undirected graph with $1,097,077$ nodes and $10,792,894$ edges, referred to as \textbf{Pokec-A} in the experiments. Using sensitive attribute \textit{gender}, we obtain an undirected graph with $1,632,803$ nodes and $22,301,964$ edges, referred to as \textbf{Pokec-G} in the experiments.
\end{itemize}

\section{Parameter Tuning}
\label{app:parameter}

The parameters in FNM include (1) $T_1$, $T_2$, $\varepsilon_1$, $\varepsilon_2$, $\mu_0$, $\xi$, and $\tau$ in Algorithm 1 as well as $T_0$, $T_3$, and $\varepsilon_3$ in Algorithm 2.
Among these parameters, the effects of $T_1$, $T_2$, $\varepsilon_1$, $\varepsilon_2$, $\tau$, $T_0$, $T_3$, and $\varepsilon_3$ on FNM are apparent: larger numbers of maximum iterations ($T_0$, $T_1$, $T_2$, $T_3$), smaller error parameters ($\varepsilon_1$, $\varepsilon_2$, $\varepsilon_3$), and smaller initial step size ($\tau$) lead to higher-quality solutions but lower time efficiency.
In practice, we keep the default values $T_0 = 100$ and $\tau = 10^{-3}$ that have been widely used in existing implementations.
For $T_1$, $T_2$, and $T_3$, we attempt to gradually increase their values until the performance of FNM cannot be improved anymore.
Then, these ``large enough'' values are used for FNM in all the remaining experiments.
The values of $\varepsilon_1$, $\varepsilon_2$, and $\varepsilon_3$ are determined similarly.

\begin{figure}[t]
  \centering
  \subfloat{\includegraphics[width=\linewidth]{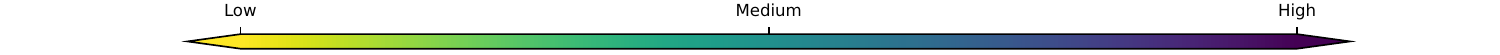}}
  \\
  \subfloat[Facebook]{\includegraphics[width=0.196\linewidth]{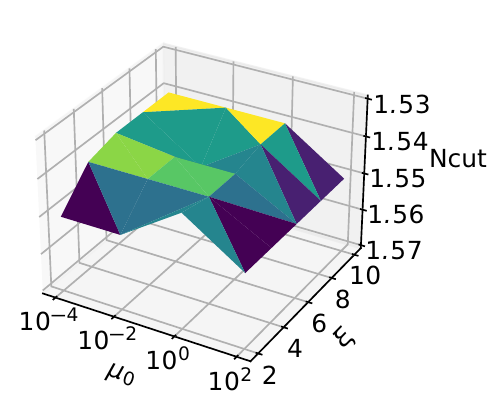}}
  \subfloat[German]{\includegraphics[width=0.196\linewidth]{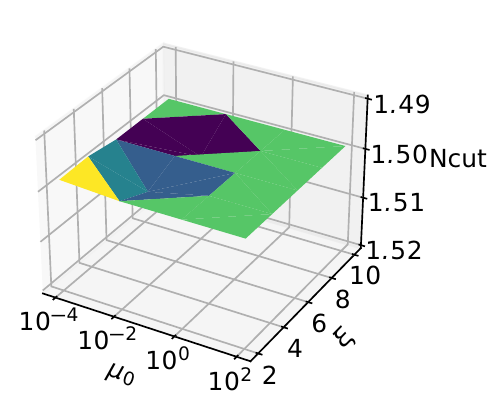}}
  \subfloat[SBM]{\includegraphics[width=0.196\linewidth]{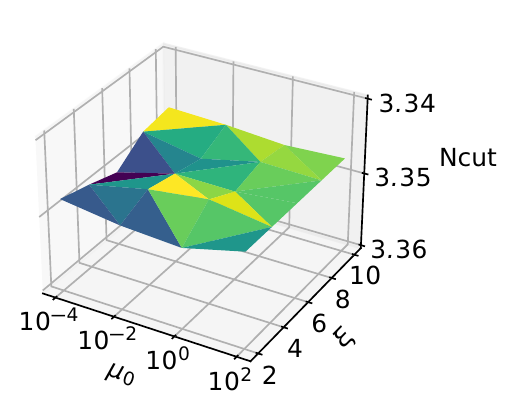}}
  \subfloat[DBLP]{\includegraphics[width=0.196\linewidth]{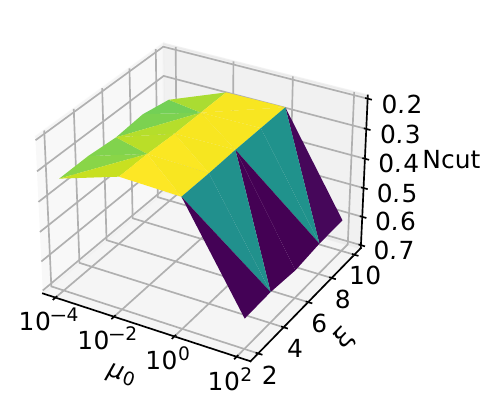}}
  \subfloat[LastFM]{\includegraphics[width=0.196\linewidth]{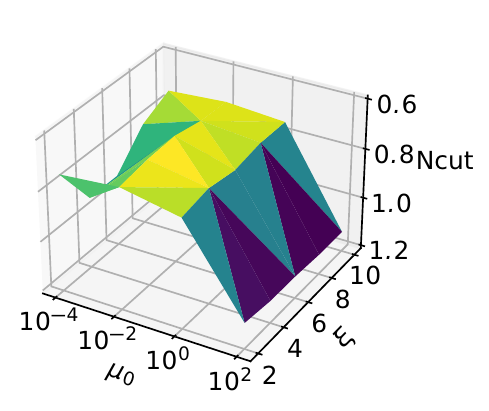}}
  \caption{Ncut values of the partitioning schemes provided by FNM for different combinations of $\xi$ and $\mu_0$.}
  \label{fig-grid}
\end{figure}

The procedure of tuning parameters $\mu_0$ and $\xi$ for FNM is a bit more complex because too large or too small values may lead to poor performance.
There are two issues to consider in the parameter tuning:
(\emph{i}) the value of $\mu_0$ should not be too large or too small to avoid the first sub-problem being ill-conditioned or the same as the unconstrained problem;
(\emph{ii}) the value of $\mu_t$ should not increase too fast or too slow over $t$ so that the number of sub-problems to solve is appropriate.
We note that several studies (e.g., \cite{curtis2015adaptive}) consider adaptively adjusting $\mu_t$ in the augmented Lagrangian method.
Nevertheless, we adopt a simple yet effective scheme that chooses an appropriate value of $\mu_0$ and uses a fixed amplification parameter $\xi > 1$ to adjust it by setting $\mu_{t+1} = \xi \mu_t$ over iterations.
In each experiment, we perform a grid search on $\xi \in \{2, 4, 6, 8, 10\}$ and $\mu_0 \in \{10^{-4}, 10^{-2}, 10^0, 10^2\}$ and select the combination of $\xi$ and $\mu_0$ with the smallest Ncut value.
The grid search procedure on five datasets when $k = 5$ and $\sigma = 0.2$ are shown in Figure~\ref{fig-grid}.
From Figure~\ref{fig-grid}, we observe that, unlike other parameters, no single combination of $\xi$ and $\mu_0$ achieves good performance across all datasets.
Therefore, we run the grid search to find appropriate values of $\xi$ and $\mu_0$ individually for each experiment.

\section{More Results on the Effects of \texorpdfstring{$\sigma$}{Sigma} and \texorpdfstring{$k$}{k}}
\label{app:additional}

\begin{figure}[t]
  \centering
  \subfloat{\includegraphics[width=\linewidth]{figures/legend-sigma.pdf}}
  \\
  \subfloat[German]{\includegraphics[width=0.196\linewidth]{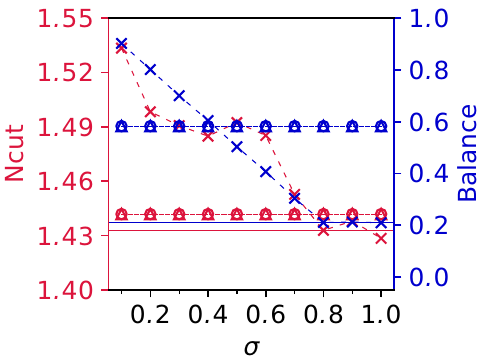}}
  \hspace{1em}
  \subfloat[Deezer]{\includegraphics[width=0.196\linewidth]{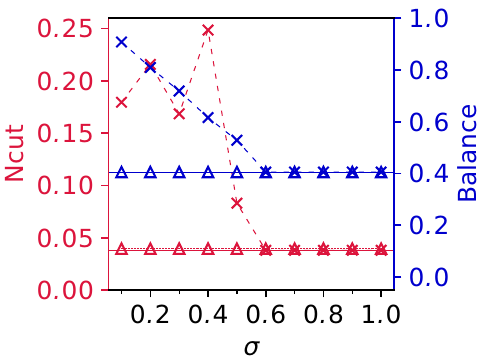}}
  \hspace{1em}
  \subfloat[Pokec-A]{\includegraphics[width=0.196\linewidth]{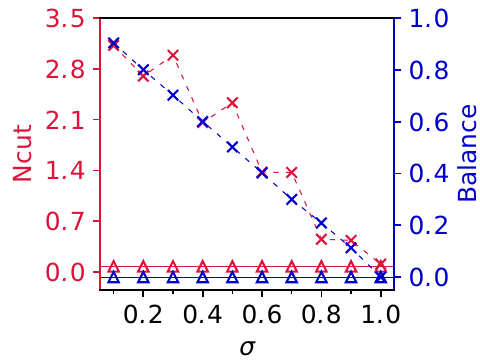}}
  \hspace{1em}
  \subfloat[Pokec-G]{\includegraphics[width=0.196\linewidth]{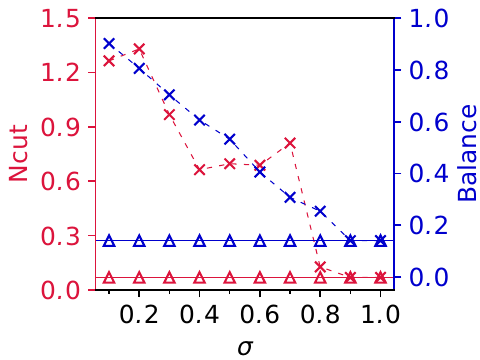}}
  \\
  \subfloat{\includegraphics[width=\linewidth]{figures/legend-time-sigma.pdf}}
  \\
  \subfloat[German]{\includegraphics[width=0.196\linewidth]{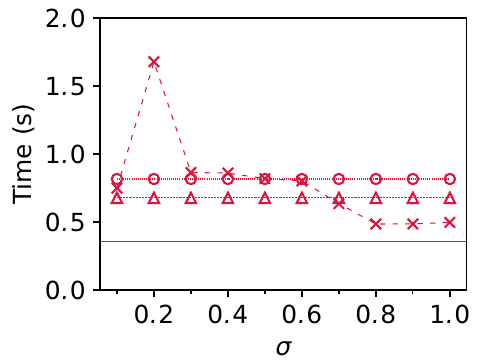}}
  \hspace{1em}
  \subfloat[Deezer]{\includegraphics[width=0.196\linewidth]{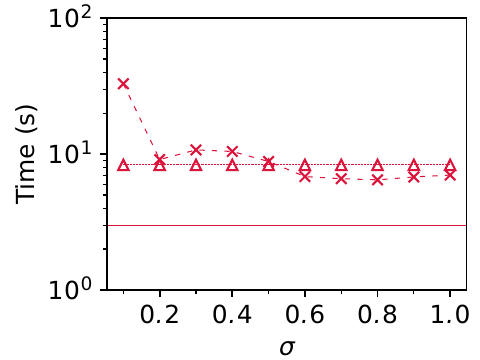}}
  \hspace{1em}
  \subfloat[Pokec-A]{\includegraphics[width=0.196\linewidth]{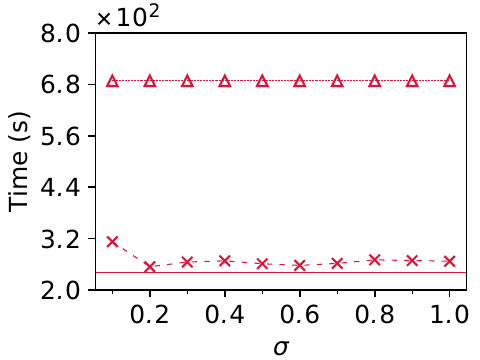}}
  \hspace{1em}
  \subfloat[Pokec-G]{\includegraphics[width=0.196\linewidth]{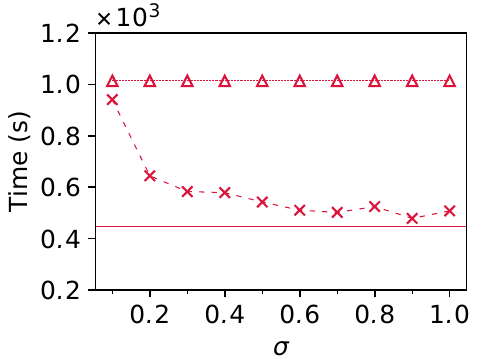}}
  \caption{Results of different algorithms with varying the fairness variable $\sigma$ on the remaining four datasets.}
  \label{fig-sigma-others}
\end{figure}

\begin{figure}[t]
  \centering
  \subfloat{\includegraphics[width=\linewidth]{figures/legend-k.pdf}}
  \\
  \subfloat[German]{\includegraphics[width=0.196\linewidth]{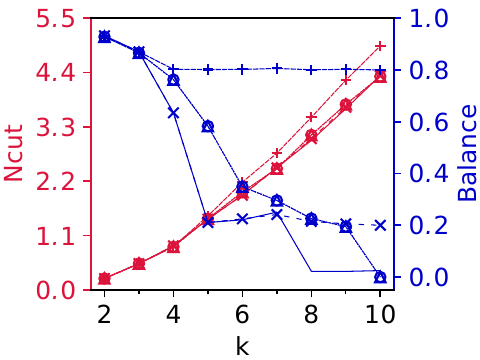}}
  \hspace{1em}
  \subfloat[Deezer]{\includegraphics[width=0.196\linewidth]{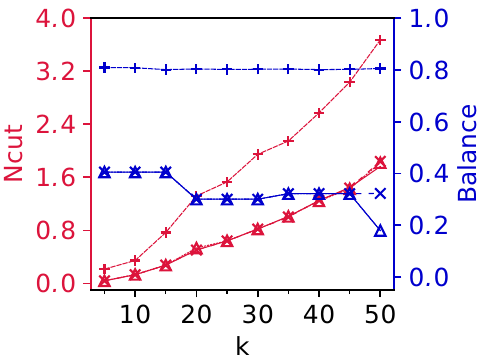}}
  \hspace{1em}
  \subfloat[Pokec-A]{\includegraphics[width=0.196\linewidth]{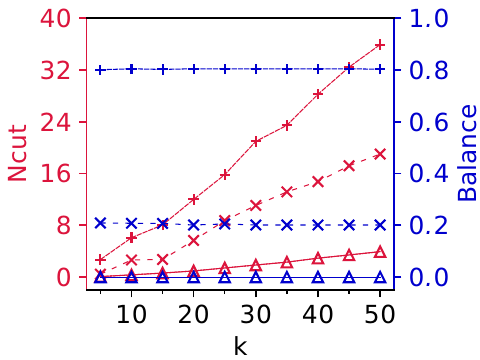}}
  \hspace{1em}
  \subfloat[Pokec-G]{\includegraphics[width=0.196\linewidth]{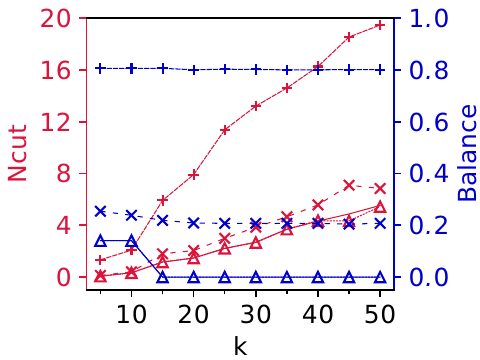}}
  \\
  \subfloat{\includegraphics[width=\linewidth]{figures/legend-time-k.pdf}}
  \\
  \subfloat[German]{\includegraphics[width=0.196\linewidth]{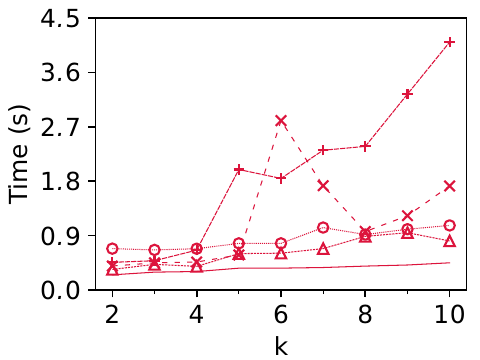}}
  \hspace{1em}
  \subfloat[Deezer]{\includegraphics[width=0.196\linewidth]{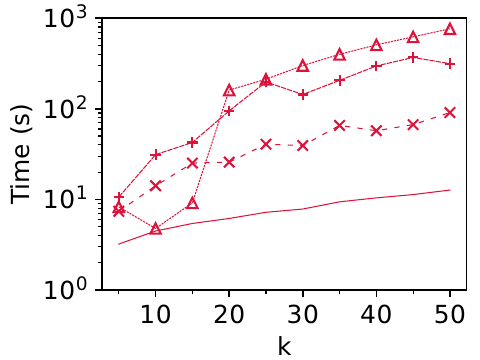}}
  \hspace{1em}
  \subfloat[Pokec-A]{\includegraphics[width=0.196\linewidth]{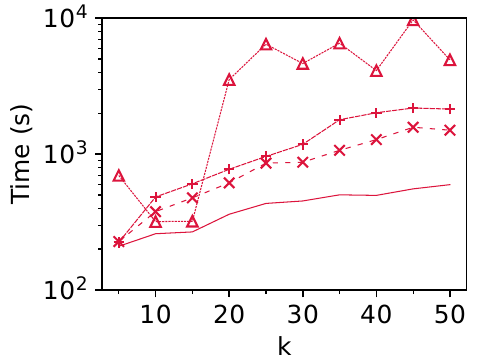}}
  \hspace{1em}
  \subfloat[Pokec-G]{\includegraphics[width=0.196\linewidth]{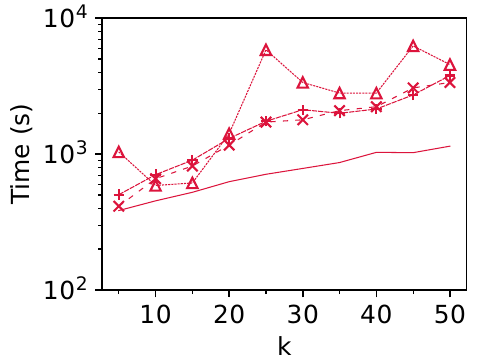}}
  \caption{Results of different algorithms as a function of $k$ on the remaining four datasets.}
  \label{fig-k-others}
\end{figure}

The first rows of Figures~\ref{fig-sigma-others} and~\ref{fig-k-others} present the results for the effects of $\sigma$ and $k$ on different algorithms in the remaining four datasets not included in Figures~\ref{fig-k} and~\ref{fig-sigma}. We observe similar trends for Ncut and balance values to those analyzed in Section~\ref{sec-exp}.

\begin{figure}[t]
  \centering
  \subfloat{\includegraphics[width=\linewidth]{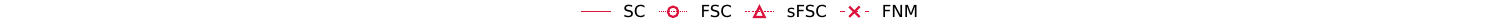}}
  \\
  \subfloat[Facebook]{\includegraphics[width=0.196\linewidth]{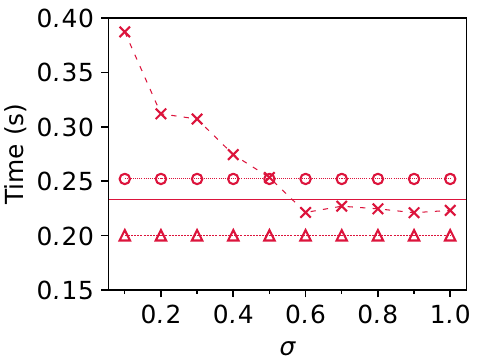}}
  \subfloat[SBM]{\includegraphics[width=0.196\linewidth]{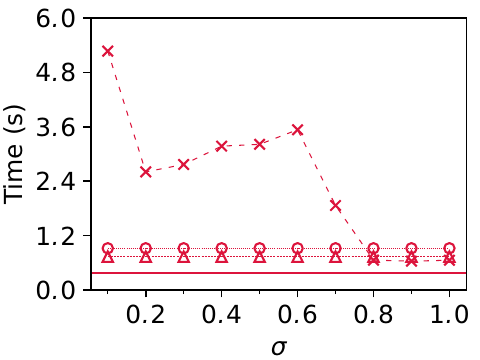}}
  \subfloat[DBLP]{\includegraphics[width=0.196\linewidth]{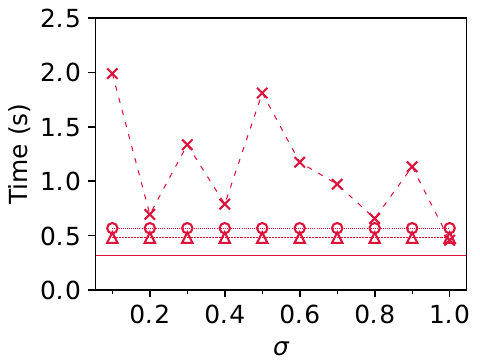}}
  \subfloat[LastFM]{\includegraphics[width=0.196\linewidth]{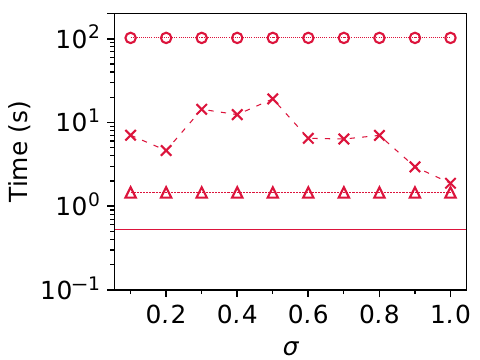}}
  \subfloat[Credit]{\includegraphics[width=0.196\linewidth]{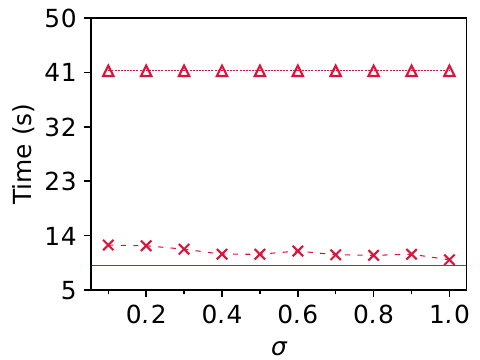}}
  \caption{Running time of different algorithms with varying the fairness variable $\sigma$ on the five datasets in Figure~\ref{fig-sigma}.}
  \label{fig-sigma-time}
\end{figure}

\begin{figure}[t]
  \centering
  \subfloat{\includegraphics[width=\linewidth]{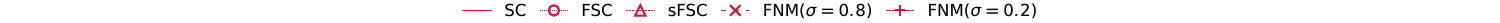}}
  \\
  \subfloat[Facebook]{\includegraphics[width=0.196\linewidth]{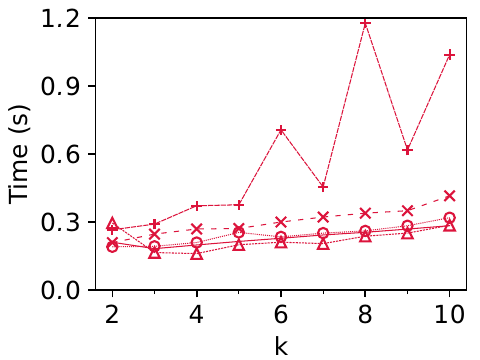}}
  \subfloat[SBM]{\includegraphics[width=0.196\linewidth]{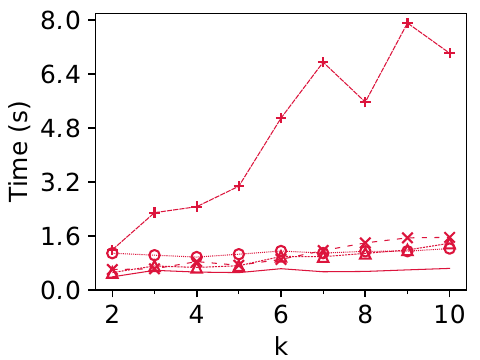}}
  \subfloat[DBLP]{\includegraphics[width=0.196\linewidth]{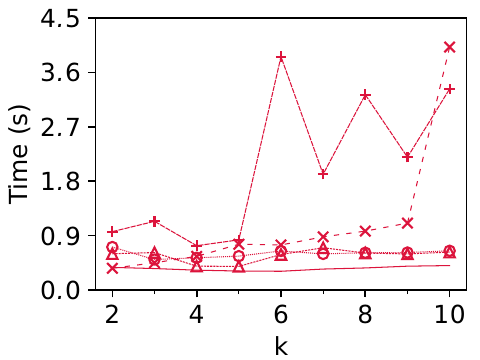}}
  \subfloat[LastFM]{\includegraphics[width=0.196\linewidth]{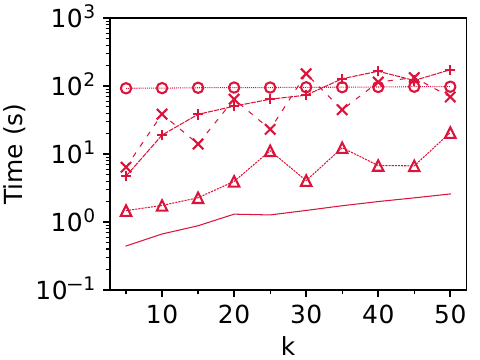}}
  \subfloat[Credit]{\includegraphics[width=0.196\linewidth]{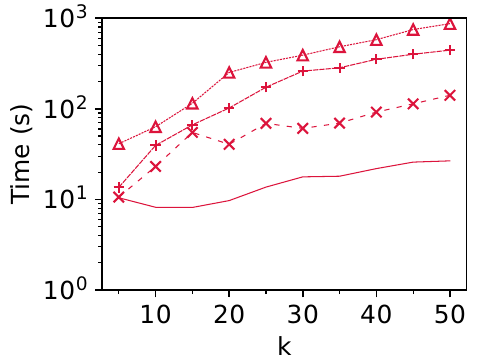}}
  \caption{Running time of different algorithms as a function of $k$ on the five datasets in Figure~\ref{fig-k}.}
  \label{fig-k-time}
\end{figure}

Next, we illustrate the running time of different algorithms with varying $\sigma$ and $k$ on all the nine datasets in Figures~\ref{fig-sigma-time} and~\ref{fig-k-time} as well as the second rows of Figures~\ref{fig-sigma-others} and~\ref{fig-k-others}.
In terms of time efficiency, we find that FNM generally runs slower when $\sigma$ is smaller or $k$ is larger.
A larger $k$ naturally leads to higher computational costs per iteration in both the embedding and rounding phases.
However, the running time of FNM is sometimes shorter when $k$ is larger because the number of iterations in Algorithm 1 is also affected by parameter settings w.r.t.~different $k$'s.
Then, a smaller $\sigma$ corresponds to a tighter fairness constraint and a more different solution from the unconstrained one, thus requiring more iterations in Algorithm 1 for convergence.
But this trend may not strictly always follow also because of various parameter settings for different values of $\sigma$.
In addition, we also observe that FNM runs slower than SC as it is more time-consuming in both the embedding and rounding phases.
But FNM runs faster than FSC in most cases since it does not require eigendecomposition.
Compared to sFSC, which improves the scalability of FSC by avoiding eigendecomposition on dense matrices, FNM runs slower on smaller graphs due to a longer time for fair rounding but becomes faster on larger graphs owing to the efficiency improvements for fair embedding.


\end{document}